\documentclass[lettersize,journal]{IEEEtran}
\usepackage{amsmath,amsfonts}
\usepackage{algorithmic}
\usepackage{algorithm}
\usepackage{array}
\usepackage[caption=false,font=normalsize,labelfont=sf,textfont=sf]{subfig}
\usepackage{textcomp}
\usepackage{stfloats}
\usepackage{url}
\usepackage{verbatim}
\usepackage{graphicx}
\usepackage{cite}
\usepackage{multirow}
\usepackage[T1]{fontenc}
\usepackage{authblk}
\begin{document}

\title{A Constrained Deformable Convolutional Network for Efficient Single Image Dynamic Scene Blind Deblurring with Spatially-Variant Motion Blur Kernels Estimation}

\author[1]{Shu Tang \thanks{
		This work was supported in part by the National Natural Science Foundation of China under Grant No. 61601070, Grant 61501074, the Key Project of Science and Technology Research of Chongqing Education Commission under Grant No. KJZD-K201800603, the Major Project of Science and Technology Research of Chongqing Education Commission under Grant No. KJZD-M201900602, the Foundation Research and Advanced Exploration Project of Chongqing under Grant No. cstc2018jcyjAX0432, the Special General Program of Technology Innovation and Application Development of Chongqing under Grant No. cstc2020jscx-msxmX0135.} \thanks{Corresponding author: tangshu@cqupt.edu.cn}}
\author[1]{Yang Wu }
\author[2]{Hongxing Qin \thanks{Corresponding author: qinhx@cqu.edu.cn}}
\author[1]{Xianzhong Xie,~\IEEEmembership{Member,~IEEE}}
\author[1]{Shuli Yang}
\author[1]{Jing Wang}

\affil[1]{Chongqing Key Laboratory of Computer Network and Communications Technology, Chongqing University of Posts and Telecommunications, Chongqing 400065, China}
\affil[2]{Chongqing University, Chongqing 400044, China}

\renewcommand\Authands{, }




\maketitle

\begin{abstract}
	
Most existing deep-learning-based single image dynamic scene blind deblurring (SIDSBD) methods usually design deep networks to directly remove the spatially-variant motion blurs from one inputted motion blurred image, without blur kernels estimation. Recently, a reblurring training strategy has been proved that it can significantly boost the deblurring performance of deep-learning-based video blind deblurring (DLVBD) methods for the motion blurred video. For the DLVBD methods, the success of reblurring training strategy mainly stems from the estimation of optical flows between two or more consecutive frames which are used to estimate/model the spatially-variant motion blur kernels of motion blurred frames and consequently guide the video deblurring. However, this strategy does not hold for the SIDSBD method as we only have one observed motion blurred image without any additional previous and next frames of the observed motion blurred image. In this paper, inspired by the Projective Motion Path Blur (PMPB) model and deformable convolution, we propose a novel constrained deformable convolutional network (CDCN) for efficient single image dynamic scene blind deblurring, which simultaneously achieves accurate spatially-variant motion blur kernels estimation and the high-quality image restoration from only one observed motion blurred image. In our proposed CDCN, we first construct a novel multi-scale multi-level multi-input multi-output (MSML-MIMO) encoder-decoder architecture for more powerful features extraction ability. Second, different from the DLVBD methods that use multiple consecutive frames, a novel constrained deformable convolution reblurring (CDCR) strategy is proposed, in which the deformable convolution is first applied to blurred features of the inputted single motion blurred image for learning the sampling points of motion blur kernel of each pixel, which is similar to the estimation of the motion density function of the camera shake in the PMPB model, and then a novel PMPB-based reblurring loss function is proposed to constrain the learned sampling points convergence, which can make the learned sampling points match with the relative motion trajectory of each pixel better and promote the accuracy of the spatially-variant motion blur kernels estimation. Extensive experiments show that our method not only can estimate spatially-variant motion blur kernels accurately, but also can produce better deblurring results than the state-of-the-art SIDSBD methods in terms of both qualitative evaluation and quantitative metrics.

\end{abstract}


\section{INTRODUCTION}
\IEEEPARstart{T}{he} blind deblurring for single motion blurred image, whose goal is to recover a clear image from its single motion blurry version, is a severe ill-posed inverse problem. Especially in the real dynamic scene, many factors such as the multidimensional relative motion between the scene and the imaging device during exposure time, the noise and depth variation make this inverse problem more challenging. To tackle such an inverse problem, numerous optimization-based methods and deep-learning-based methods have been developed to model the blur process and regularize the solution space, and learn the mapping function between the clear and blurry image pairs, respectively. 

For the optimization-based methods, the key to success is to build the right model to model the formation process of image blur: the convolution operation for the uniform motion blur\cite{pan2016blind,pan2016l_0,tang2018spatial,wang2011analyzing,xu2013unnatural}, the efficient filter flow and the Projective Motion Path Blur (PMPB) model for the spatially-variant motion blur\cite{gupta2010single,tai2010richardson,harmeling2010space,hirsch2011fast,whyte2012non,hu2014joint,sheng2019depth}, and so on. Among these models, the PMPB model, which takes a motion blurred image as the result of integrating all intermediate images the camera “sees” along the trajectory of the relative motion, has been proved to be one of the best motion blur models that can model spatially-variant motion blurs well . However, in the optimization-based methods, the PMPB model requires computing all possible combinations of all motion spaces, which inevitably leads to tremendous computational cost and hence can only model small-size 3-dimensional camera shake.

Recently, the deep-learning-based methods have achieved significant improvement in single image dynamic scene blind deblurring (SIDSBD). Most existing deep-learning-based SIDSBD methods focus on learning the regression relation between a motion blurry input image and the corresponding clear image in an end-to-end manner, which skip the estimation of motion blur kernels\cite{ramakrishnan2017deep,nah2017deep,kupyn2018deblurgan,tao2018scale,zhang2018dynamic,zhang2019deep,gao2019dynamic,cai2020dark,yuan2020efficient,zamir2021multi,cho2021rethinking,purohit2021spatially,wang2022uformer,zamir2022restormer,wan2020deep,ulyanov2018deep,zhang2020deblurring,park2020multi,suin2020spatially}. Nevertheless, in recent years, a reblurring training strategy has been widely applied to the deep-learning-based video blind deblurring (DLVBD) methods, which can significantly boost the deblurring performance of DLVBD methods\cite{chen2018reblur2deblur,zhang2021deep,bai2022self,wang2022MMP}. The success of the reblurring training strategy mainly stems from the estimation or modeling of the spatially-variant motion blur kernels of each motion blurred frame, which is achieved by using the optical flows between two or more consecutive frames in the motion blurred video. These phenomenon set us thinking that whether the estimation of spatially-variant motion blur kernels and the corresponding reblurring training strategy are still good for the SIDSBD? And obviously, the strategy used by the DLVBD method does not hold for the SIDSBD method as we only have one observed motion blurred image without any additional previous and next frames of the observed motion blurred image. Therefore, the innovative approaches, which can dig out informative blurred features and consequently achieve accurate spatially-variant motion blur kernels estimation from a single motion blurred image, need to be explored.

In this paper, we propose a novel constrained deformable convolutional network (CDCN) for accurate spatially-variant motion blur kernels estimation and higher quality image restoration from only one observed motion blurred image. Specifically, inspired by the PMPB model and the deformable convolution ,we propose a novel constrained deformable convolution reblurring (CDCR) strategy, in which the deformable convolution is first used to learn the sampling points of spatially-variant motion blur kernels of the inputted single motion blurred image, which is similar to the estimation of the motion density function of the camera shake in the PMPB model, and then a novel PMPB-based reblurring loss function is proposed to constrain the learned sampling points convergence, which optimizes the estimation of the spatially-variant motion blur kernels. To the best of our knowledge, our CDCN is the first deep-learning-based SIDSBD method that can accurately estimate spatially-variant motion blur kernels from only one single motion blurred image without the optical flow. Our proposed network can be trained in an end-to-end manner. In summary, the main contributions of our proposed CDCN are listed as follows:

1) We propose a CDCR strategy, which achieves accurate spatially-variant motion blur kernels estimation from only one single motion blurred image without the optical flow by constraining the convergence of the sampling points of spatially-variant motion blur kernels with the PMPB-based reblurring loss function.  

2) A small convolutional neural network (CNN) with one SoftMax layer, several PReLU layers and several convolutional layers is constructed for predicting the inverse kernel of each estimated motion blur kernel. The predicted inverse kernels are directly applied to blurred features of the inputted single motion blurred image to generate deblurred features, which can enhance the restoration ability of the decoder and achieve better deblurring performance for SIDSBD.

3) We construct a novel multi-scale multi-level multi-input multi-output (MSML-MIMO) encoder-decoder architecture via combining the multi-input multi-output strategy into our early research work (i.e. the multi-scale channel attention network: MSCAN\cite{wan2020deep}), which can further enhance features extraction ability of the network and consequently facilitate accurate estimation of spatially-variant motion blur kernels and higher quality image restoration.

\section{RELATED WORK}

Numerous optimization-based and deep-learning-based deblurring methods have been proposed in the literatures. Due to the space limitation, here we focus on works related to our method.

\subsection{The Optimization-based Approach}

Agrawal et al.\cite{agrawal2009invertible} pointed out that the reason why deconvolution is a highly ill-posed problem is that there are many null values in the point spread function (PSF) frequency domain space. By changing the exposure time of each frame in the video, a series of frames were used to fill the null value in the PSF frequency domain space of the blurred frame, so that the motion blur of the blurred frame became a reversible process, so as to solve the problem of deblurring objects moving at a uniform speed.
Many researchers thought that an image blurred by camera shake can be viewed as the result of integrating all intermediate images the camera “sees” along the trajectory of the camera shake, therefore the so-called projective motion path blur (PMPB) model was proposed to model the spatially-variant blur. Gupta et al.\cite{gupta2010single} proposed a PMPB-based motion density function (MDF) to describe the exposure time spent on the three-dimensional motion trajectory of the camera, which was used to estimate the spatially-variant motion blur kernels.
Tai et al.\cite{tai2010richardson} regarded the blurred image as the  integration of a series of clear scenes, which went through a sequence of planar projective transformations, and proposed a PMPB-based RL algorithm, which can incorporate many regularization priors to improve the deblurred results.
Harmeling et al.\cite{harmeling2010space} proposed a space-variant blind deblurring method based on filter flow by studying the type of camera jitter, and designed an experimental device to record the space-variant PSF corresponding to the blur while taking the blurred image.
Hirsch et al.\cite{hirsch2011fast} combined with the PMPB model and the Efficient Filter Flow (EFF), and proposed an efficient algorithm that can deal with non-uniform blur caused by camera shake.
Whyte et al.\cite{whyte2012non} assumed that the camera rotation was the only significant source of camera shake blur, and proposed a PMPB-based parameterized geometric model to remove the non-uniform camera rotation blur. 
Xu et al.\cite{xu2012depth} proposed a hierarchical estimation framework based on region trees to estimate the blur kernel step by step, and redesigned a spatially varying PSF estimation algorithm based on shock filtering invariance for non-uniform deblurring.
Hu et al.\cite{hu2014joint} thought that non-uniform blur was not only caused by camera shake, but also by the change of scene depth. Therefore, a method of simultaneously estimating scene depth and removing non-uniform blur was proposed.
Sheng et al.\cite{sheng2019depth} proposed a PMPB-based depth-aware motion blur model with a given depth image. The authors used a PatchMatch-based depth filling method to fix the empty holes in the depth image. The Deblurring and depth filling were performed iteratively to refine the results. 
Bai et al.\cite{bai2019single}  observed that a coarse enough image down-sampled from a blurry observation was very close to a low-resolution version of the latent sharp image. Based on this observation, the authors proposed a coarse-to-fine progressive single-image blind deblurring algorithm.
Ulyanov et al.\cite{ulyanov2018deep} showed that an elaborate UNet is sufficient to capture the statistics prior of a single image for the low-level tasks learning.
Inspired by\cite{ulyanov2018deep}, Ren et al.\cite{ren2020neural} proposed a self-supervised blind deblurring method for a single uniform blurred image, which combined deep models with the maximum a posterior (MAP), and constructed two generative networks for the latent image restoration and the blur kernel estimation, respectively. 

From above discussions we can see that most optimization-based methods\cite{agrawal2009invertible,gupta2010single,tai2010richardson,harmeling2010space,hirsch2011fast,whyte2012non,xu2012depth,hu2014joint,sheng2019depth,bai2019single,ren2020neural} can only handle either small-size camera shake without the movement of objects or the rigid object motion without camera shake. Although the PMPB model can model spatially-variant motion blurs very well, it suffers from tremendous computational cost and hence can only model small-size 3-dimensional camera shake\cite{gupta2010single,tai2010richardson,harmeling2010space,whyte2012non,hu2014joint,sheng2019depth}. So, the optimization based methods are not suitable for the real-word complex dynamic scene deblurring problem, which contains camera shake, multiple rigid or non-rigid objects motion, and different scene depths simultaneously.

\subsection{The Deep-Learning-Based Approach}

Lately, impressive progress has been made in SIDSBD by using deep-learning-based single image blind deblurring methods. Xu et al.\cite{xu2014deep} proposed an end-to-end CNN-based non-blind deblurring network to learn the deconvolution operation for the disk and motion blurs. 
Ramakrishnan et al.\cite{ramakrishnan2017deep} proposed a generative advantageous network (GAN) whose generator consisted of the global jump links and dense connections. Ramakrishnan et al. obtained the restoration image from the inputted blurred image directly, and without the blur kernel estimation.
Nah et al.\cite{nah2017deep} designed a multi-scale end-to-end deblurring network,and proposed an improved residual block for SIDSBD. Again, there was no blur kernel estimation.
The DeblurGAN model proposed by kupyn et al.\cite{kupyn2018deblurgan} greatly improved the values of the self-similarity measure (SSIM) and subjective visual effects through well-designed advantageous loss and content loss.
Tao et al.\cite{tao2018scale} proposed a scale recurrent network (SRN), which applied the ResBlock to the encoder-decoder module, and restored sharp images with different resolutions gradually.
Zhang et al.\cite{zhang2018dynamic} proposed a spatially-variant recurrent neural network (RNN) for spatially-variant blurs, in which different weights were learned for different pixels.
Inspired by the spatial pyramid matching (SPM), Zhang et al.\cite{zhang2019deep} proposed a deep multi-patch hierarchical network (DMPHN) to achieve end-to-end non-uniform deblurring. The DMPHN made inputs at different levels have the same spatial resolution, therefore, the residual manner could be introduced between levels.
Gao et al.\cite{gao2019dynamic} proposed a nested skip connection structure to replace the conventional residual connection, and a parameter selection sharing strategy between different scales for SIDSBD.
Cai et al.\cite{cai2020dark} thought that appropriate image priors and regularization terms could improve the deblurring performance. Therefore, they inserted an extreme channel prior into the CNN-based blind deblurring network, and proposed an extreme channel prior embedded network (ECPeNet) for the SIDSBD.
Yuan et al.\cite{yuan2020efficient} introduced the blur kernel estimation into deep-learning-based SIDSBD, and proposed a spatially variant deconvolution network (SVDN). In their proposed SVDN, the deformable convolution was first used to learn the sampling points of the blur kernels, and then the optical flows between the blurred image and it's nearby frames were used to guide the learning of the sampling points.
Zamir et al.\cite{zamir2021multi} combined the characteristics of the encoder-decoder network with the single scale network, and designed a multi-stage network structure, in which not only the attention mechanism was introduced into each stage of the network, but also an information exchange strategy between different stages was proposed.
Cho et al.\cite{cho2021rethinking} proposed a multi-input multi-output encoder-decoder structure and an asymmetric feature fusion strategy  to fuse multi-scale features effectively.
Purohit et al.\cite{purohit2021spatially} thought that different regions of the blurred image had different degrees of degradation, so they first designed a positioning network to identify the degraded regions, and then the learned degradation features were used to guide the recovery network for adaptive deblurring.
Wang et al.\cite{wang2022uformer} combined the UNet and the Transformer, and proposed a general u-shaped transformer for image restoration. In their proposed network,  a block-based self-attention and a learnable multi-scale recovery modulator were proposed to capture the local features and estimate the modulation parameter of each window, respectively.
In order to reduce the computational overhead of the Transformer on low-level visual tasks, Zamir et al.\cite{zamir2022restormer} proposed a multi-dconv head transposed attention (MDTA) module, which calculated attention in the channel dimension rather than the pixel dimension. In addition, a gated-dconv feed-forward network (GDFN) was proposed to capture the local information of images

Except for the SIDSBD, recently, the deep-learning-based video blind deblurring (DLVBD) methods have also been greatly developed and achieved significant improvement in video motion blind deblurring. Su et al.\cite{su2017deep} proposed an end-to-end video deblurring network based on the encoder-decoder  architecture, and collected real-world motion blurred video datasets using high frame rate cameras.
Chen et al.\cite{chen2018reblur2deblur} used a optical flow network to estimate the optical flows between  consecutive frames, and the estimated optical flows were used to estimate the spatially-variant motion blur kernels. The estimated blur kernels were used to fine-tune the proposed deblurring network by using a reblurring self-supervised loss, which could boost the deblurring performance of the proposed deblurring network. 
Zhang et al.\cite{zhang2021deep} proved that, in a motion blurred video, the pixel-wise blur kernel can be represented by the pixel-wise optical flow. Therefore, Zhang et al. used the optical flows to model spatially-variant motion blur kernels, which were used to generate the weights of the RNNs for video blind deblurring.
Bai et al.\cite{bai2022self} estimated the blur kernel form a low-resolution and uniform blurred video directly, and used the estimated blur kernel to achieve self-supervised-based high-resolution video restoration.
Wang et al.\cite{wang2022MMP} first estimated the average magnitude of optical flows from several clear consecutive frames, then the estimated average magnitude was used to learn the pixel-wise motion blur level of each motion blurred frame. Then, Wang et al. utilized the learned motion blur levels as guidance for effective deep video deblurring.  

From above discussions we can see that, on the one hand, most existing deep-learning-based single image dynamic scene blind deblurring (SIDSBD) methods usually design deep networks to directly remove the spatially-variant motion blurs from one inputted motion blurred image, without blur kernels estimation\cite{ramakrishnan2017deep,nah2017deep,kupyn2018deblurgan,tao2018scale,zhang2018dynamic,zhang2019deep,gao2019dynamic,cai2020dark,yuan2020efficient,zamir2021multi,cho2021rethinking,purohit2021spatially,wang2022uformer,zamir2022restormer}. On the other hand, the estimation of the spatially-variant motion blur kernels and the corresponding reblurring training strategy have been widely applied to the deep-learning-based video blind deblurring (DLVBD) methods and have proved their effectiveness\cite{chen2018reblur2deblur,zhang2021deep,bai2022self,wang2022MMP}. Therefore we wonder that whether the estimation of spatially-variant motion blur kernels and the corresponding reblurring training strategy are still good for the SIDSBD? So, in this paper, we mainly focus on the research of the estimation of the spatially-variant motion blur kernels and the corresponding reblurring training strategy for SIDSBD.

\section{THE PROPOSED CDCN}
\begin{figure*}[!t]
	\centering
	\includegraphics[width=6.5in]{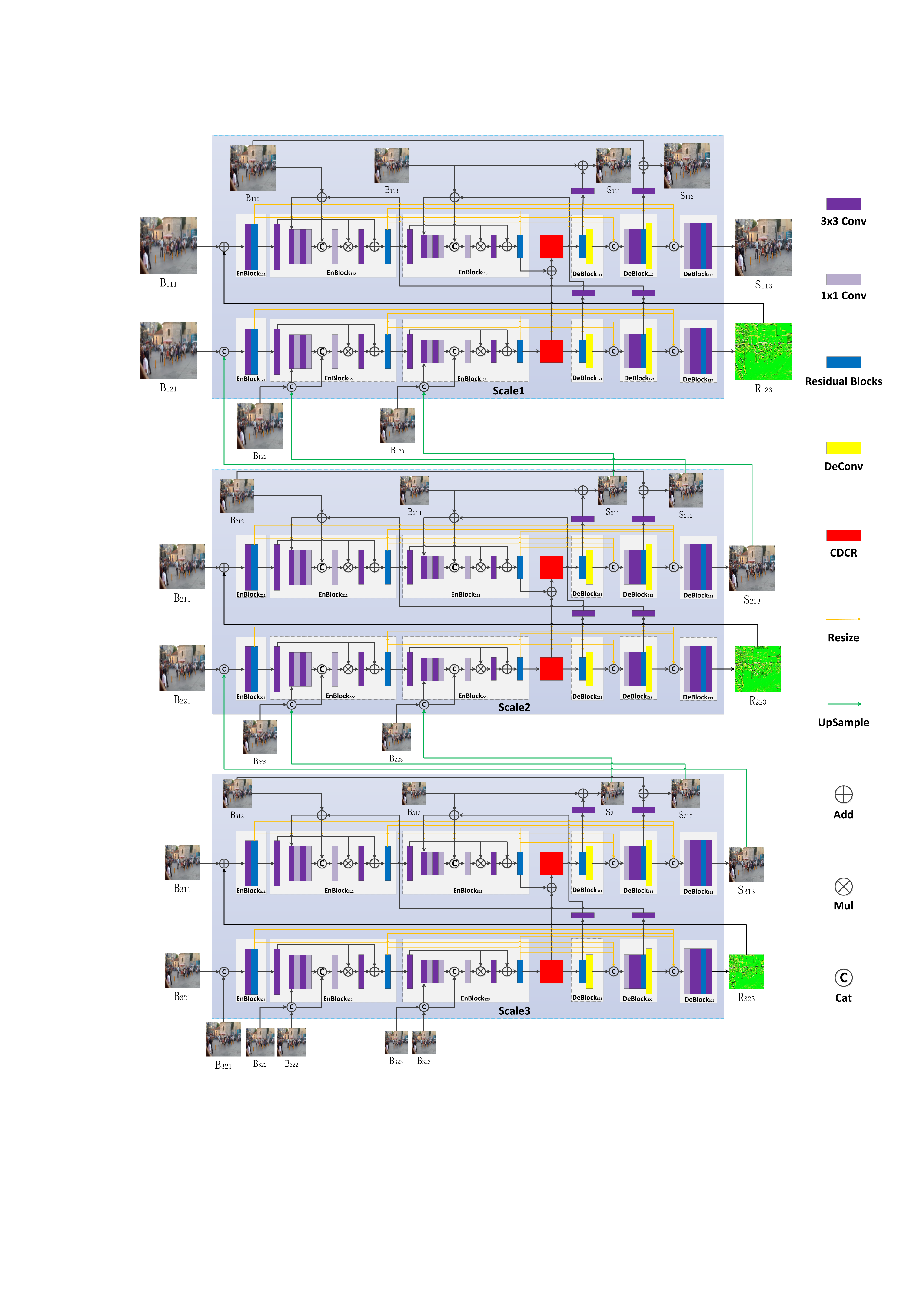}
	\caption{Our proposed constrained deformable convolutional network (CDCN).}
	\label{MSML_MIMO}
\end{figure*}

Our proposed CDCN for SIDSBD is illustrated in Fig. \ref{MSML_MIMO}. As we discussed above, the main contributions of our proposed CDCN are a novel MSML-MIMO encoder-decoder architecture for more powerful features extraction ability and a CDCR strategy for accurate spatially-variant motion blur kernels estimation and the corresponding inverse kernels prediction. Therefore, in this section, we will first discuss the MSML-MIMO encoder-decoder architecture and then the CDCR strategy in detail.

\subsection{The MSML-MIMO Encoder-Decoder Architecture}

Our proposed MSML-MIMO encoder-decoder architecture is illustrated in Fig. \ref{MSML_MIMO},where $B_{ijk},S_{i1k},R_{i2k}(i\in{\{1,2,3\}},j\in{\{1,2\}},k\in{\{1,2,3\}})$ represent the blur image, the recoverd image and the residual image at different scales, different levels, and different orders, respectively. In detail, $B_{ijk}, S_{i1k}, R_{i2k}(i\in{\{1,2,3\}},j\in{\{1,2\}},k\in{\{1,2,3\}})$ denote the observed blurred image input into the $k-th$ encoder block (EnBlock) at the $j-th$ level of the $i-th$ scale, the restoration image inferred by the $k-th$ decoder block (DeBlock) at the 1st level of the $i-th$ scale, and the residual image generated by the $k-th$ DeBlock at the 2nd level of the $i-th$ scale, respectively. $B_{1jk},B_{2jk}$ and $B_{3jk}$ are a sequence of blurry images downsampled from the observed original full resolution blurred image at different scales. $EB_{ijk}$, $DB_{ijk}$ and $CDCR_{ij}$ denote the $k-th$ EnBlock at the $j-th$ level of the $i-th$ scale, the $k-th$ DeBlock at the $j-th$ level of the $i-th$ scale, and the CDCR module at the $j-th$ level of the $i-th$ scale, respectively. 

As shown in the Fig. \ref{MSML_MIMO}, our proposed MSML-MIMO encoder-decoder architecture includes three scales, in which each scale consists of two levels, and each level contains three EnBlocks, three DeBlocks, and a CDCR module. Except the $EB_{ij1}$, the first convolution of $EB_{ij2}$ and $EB_{ij3}$ is a strided convolution with a stride of $2$ for downsampling. And the stride of the transposed convolution in $DB_{ij1}$ and $DB_{ij2}$ is $2$ for upsampling. For the $DB_{i21}$ and $DB_{i22}$, two additional $3\times 3$ convolution layers with the output channel of $3$ are used to obtain the intermediate residual images with different scales respectively. And, for the $DB_{i11}$ and $DB_{i12}$, two additional $3\times 3$ convolution layers with the output channel of $3$ are used to obtain the intermediate restoration images with different scales respectively.

As illustrated in Fig. \ref{MSML_MIMO}, the deblurring process of CDCN starts at the second/down level of the third scale. At the second level of each scale, the inputs of $EB_{321}$, $EB_{221}$ and $EB_{121}$ are $EB_{321}^{in}=B_{321}\copyright B_{321}$, $EB_{221}^{in}=B_{221}\copyright S_{313}^{\uparrow}$ and $EB_{121}^{in}=B_{121}\copyright S_{213}^{\uparrow}$, respectively. While, the input of $EB_{32k}$ includes the output of the $EB_{32(k-1)}$ and the $B_{32k}(k\in{\{2,3\}})$, and the input of $EB_{i2k}$ includes the output of the $EB_{i2(k-1)}$ and the concatenation of the $B_{i2k}$ and the $S_{(i+1)1(4-k)}^{\uparrow} (i\in{\{1,2\}},k\in{\{2,3\}})$. At the first/up level of each scale, the input of $EB_{i11}$ is $EB_{i11}^{in}=B_{i11}\oplus R_{i23}$. While, the input of $EB_{i1k}$ includes the output of the $EB_{i1(k-1)}$ and the addition of the $B_{i1k}$ and the $R_{i2(4-k)} (i\in{\{1,2\}},k\in{\{2,3\}})$. Where $EB_{ijk}^{in}$ and $S_{ijk}^{\uparrow}$ denote the input of the $EB_{ijk}$ and the upsampled version of the $S_{ijk}$ (the green arrow in Fig. \ref{MSML_MIMO}) respectively.  

For the DeBlocks, the input of the $DB_{ij1}$ is the $CDCR_{ij}$. And the input of the $DB_{ijk} (k\in{\{2,3\}})$ is the concatenation of the $DB_{ij(k-1)}^{out}$, $EB_{ij1}^{out}$, $EB_{ij2}^{out}$, and $EB_{ij3}^{out}$, by using the resize operation (the earthy$ $ yellow arrow in Fig. \ref{MSML_MIMO}) which can be formulated as:
\begin{equation}
	\label{resize1}
	DB_{ij2}^{in}=DB_{ij1}^{out}\copyright EB_{ij1}^{out\downarrow}\copyright EB_{ij2}^{out}\copyright EB_{ij3}^{out\uparrow}
\end{equation}
\begin{equation}
	\label{resize2}
	DB_{ij3}^{in}=DB_{ij2}^{out}\copyright EB_{ij1}^{out}\copyright EB_{ij2}^{out\uparrow}\copyright EB_{ij3}^{out\uparrow}
\end{equation}

where,$ EB_{ijk}^{out}$ and $DE_{ijk}^{out}$ denote the outputs of the $EB_{ijk}$ and $DE_{ijk}$ respectively. The CDCR module is used to estimate the spatially-variant motion blur kernels and predict the corresponding inverse kernels and output the deblurred features. At the second level of each scale, the input of the $CDCR_{i2}$ is the $EB_{i23}^{out}$. And At the first/up level of each scale, the input of the $CDCR_{i1}$ is $CDCR_{i1}^{in}=EB_{i13}^{out}\oplus CDCR_{i2}^{out}$, where $CDCR_{ij}^{in}$ and $CDCR_{ij}^{out}$ denote the input of the $CDCR_{ij}$ module and the output of the $CDCR_{ij}$ module respectively.

From above analyses and Fig. \ref{MSML_MIMO} we can see that, first, our MSML-MIMO encoder-decoder architecture can conduct the residual between levels. However, beyond our early work in \cite{wan2020deep}, our proposed MSML-MIMO encoder-decoder architecture conducts the residual manner between levels not only once but three times, which are between three DeBlocks of the second level and the corresponding three EnBlocks of the first level within the same scale. Second, similar to the rich residual connects between levels, for each scale, we conduct the intermediate supervision not only once but apply multiple intermediate supervisions to all DeBlocks of the first level of each scale. And except for the scale $1$, all intermediate supervision results in previous scale will be used to guide the image restoration at this scale. Finally, different from most conventional coarse-to-fine image blind deblurring networks, which fused different depths of features only between EnBlocks and DeBlocks with the same spatial resolution, our proposed MSML-MIMO encoder-decoder architecture fuses features from different spatial resolutions within each level: the second and third DeBlocks of each level take the outputs of all EnBlocks within the same level as the inputs and merge different resolution features using the concatenation operation and convolutional layers. Therefore, because of the utilization and fusion of more information flows and informative features, our proposed CDCN possesses more powerful features extraction ability, which can facilitate accurate estimation of spatially-variant motion blur kernels and higher quality image restoration for SIDSBD.. 

\subsection{THE CDCR STRATEGY}

\begin{figure}[!t]
	\centering
	\includegraphics [width=3.5in]{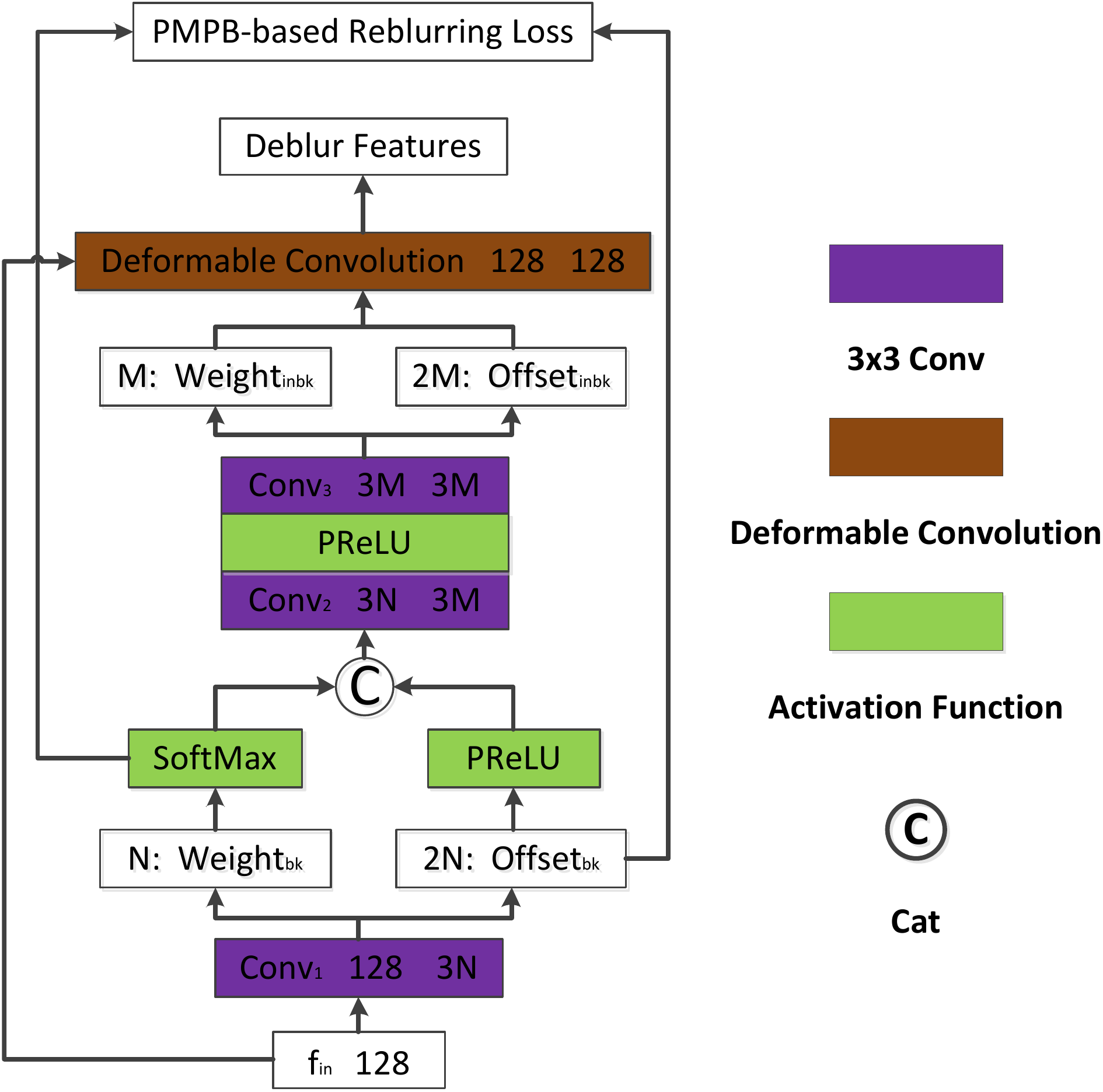}
	\caption{Our proposed constrained deformable convolution reblurring (CDCR) strategy.}
	\label{CDCR}
\end{figure}

As illustrated in Fig. \ref{CDCR}, Our proposed CDCR strategy consists of a CDCR module and a PMPB-based reblurring loss function, which is responsible for learning the sampling points of spatially-variant motion blur kernels and predicting the sampling points of their corresponding inverse kernels from one single motion blurred image. In this subsection, we will discuss the CDCR module and the PMPB-based reblurring loss function in detail. 

As shown in Fig. \ref{CDCR}, the proposed CDCR module is a small CNN, which consists of one SoftMax layer, two PReLU layers, three regular convolutional layers and one deformable convolution. On the one hand, the deformable convolution has been proved that it can make the spatial sampling locations focus on the interested image content efficiently\cite{dai2017deformable,zhu2019deformable}. On the other hand, the PMPB model, which takes a motion blurred image as the result of integrating all intermediate images the camera “sees” along the trajectory of the relative motion, is one of the best motion blur models for modeling spatially-variant motion blurs. Therefore, inspired by the deformable convolution and the PMPB model, in our proposed CDCR module, we first use a regular convolution to learn the spatial sampling locations and the corresponding weights of the trajectory of the relative motion for each pixel, which is similar to the estimation of the motion density function of the camera shake in the PMPB model. The learning of the spatial sampling locations and the corresponding weights can be formulated as:
\begin{equation}
	\label{offset1}
	offset_{bk},weight_{bk}=Sep(Conv_1(f_{in}))
\end{equation}
\begin{equation}
	\label{weight1}
	weight_{bk}=SoftMax(weight_{bk})
\end{equation}
where,$f_{in}$ denotes the input features. For the $CDCR_{i2}$, $f_{in}$ is the blurred features extracted by the EnBlocks, i.e. the $EB_{i23}^{out}$. For the $CDCR_{i1}$,$f_{in}=EB_{i13}^{out}+CDCR_{i2}^{out}$. $Conv_1()$ denotes the regular convolution operation with $3N$ output channels, where $N$ denotes the number of the sampling points of the blur kernel. $Sep()$ denotes the separation operation, which divides $3N$ channels into $2N$ and $N$ for $offset_{bk}$ and $weight_{bk}$, respectively. And $offset_{bk}=[offset_{bk,1},offset_{bk,2}...offset_{bk,N}],offset_{bk,n}=(offset_{bk,n,x},offset_{bk,n,y}),n\in{\{1,2,...N\}},$ where $offset_{bk,n,x}$ and $offset_{bk,n,y}$ denote the $x$-coordinate and the $y$-coordinate of the $n-th$ sampling point of the motion blur kernel for each pixel, respectively. 
Therefore, $offset_{bk}$ is the spatial locations of $N$ sampling points, which approximates the trajectory of the relative motion of each pixel (i.e. the shape of the motion blur kernel). And $weight_{bk}=[weight_{bk,1},weight_{bk,2}...weight_{bk,N}]$, where $weight_{bk,n}$ denotes the fraction of the exposure time (FET) spent at $n-th$ sampling point of the motion blur kernel for each pixel. Based on the principle of the PMPB model, the $SoftMax$ operation is applied to the $weight_{bk}$ for energy conservation. Then, we propose a PMPB-based reblurring loss function to constrain the accuracy of the $offset_{bk}$ and $weight_{bk}$:
\begin{equation}
	\label{ReBlur Loss}
	L_{reblur}=\sum_{i=1}^{3} \sum_{j=1}^{2}(\frac{1}{M}\sum_{m=1}^{M} \vert \vert (B_{reblur_{ij3}})^m-(B_{ij3})^m\vert \vert_2^2)
\end{equation}
where, $(B_{ij3})^m$ denotes the $m-th$ observed blurred image at the $3-th$ EnBlock of the $j-th$ level of the $i-th$ scale and $(B_{reblur_{ij3}})^m=\sum_{n=1}^{N}warp((offset_{bk,n})_{ij3}^m, (S_{GT_{i11}})^m)\bigotimes  (weight_{bk,n})_{ij3}^m$ denotes the $m-th$ reconstructed blurred image using the PMPB model at the $3-th$ EnBlock of the $j-th$ level of the $i-th$ scale. And, $(S_{GT_{i11}})^m$ is the $m-th$ ground truth sharp image at the $1-th$ DeBlock of the $1-th$ level of the $i-th$ scale. $(offset_{bk,n})_{ij3}^m$ and $(weight_{bk,n})_{ij3}^m$ denote the learned $offset_{bk,n}$ and $weight_{bk,n}$  for $(B_{ij3})^m$. 
$\vert \vert \quad \vert \vert_2^2$ and $\bigotimes$  denote the $L2$ norm and the element-wise multiplication respectively, and $warp()$ denotes the warp operation by using the bilinear interpolation. 

From Equations (\ref{ReBlur Loss}) we can see that, our proposed PMPB-based reblurring loss function can constrain the solution spaces of the learned $offset_{bk}$ and $weight_{bk}$, which makes the learned sampling points fit the trajectory of the relative motion of each pixel well and achieves accurate spatially-variant motion blur kernels estimation from a single motion blurred image.

Then, after the learning of spatially-variant motion blur kernels, we apply two PReLU layers and two regular convolutional layers to the $offset_{bk}$ and $weight_{bk}$ for predicting the inverse kernels, which can be formulated as:
\begin{equation}
	\label{Conv2 out}
	Conv_2^{out}=Conv_2(Pre(offset_{bk})\copyright weight_{bk})
\end{equation}
\begin{equation}
	\label{offset2 weight2}
	offset_{inbk},weight_{inbk}=Sep(Conv_3(Pre(Conv_2^{out})))
\end{equation}
where, $Pre()$ denotes the PReLU activation function and $\copyright$ denotes the concatenation operation. $Conv_3()$ denotes the regular convolution operation with $3M$ output channels, where $M$ denotes the number of the sampling points for the inverse kernel. Again, $Sep()$  divides $3M$ channels into $2M$ and $M$ for $offset_{inbk}$ and $weight_{inbk}$, respectively. Similar to $offset_{bk}$, $offset_{inbk}$ is the spatial locations of $M$ sampling points of the inverse kernel for each pixel, which approximates the shape of the inverse kernel. And similar to $weight_{bk}$, $weight_{inbk}$ is the weights of $M$ sampling points of the inverse kernel for each pixel but without the $SoftMax$ operation.

Finally, the predicted inverse kernels are directly applied to $f_{in}$ for generating the deblurred features by using one deformable convolution.

\subsection{THE LOSS FUNCTION}

The loss function of our CDCN consists of the multi-scale content mean square error loss, the multi-scale frequency reconstruction loss and the PMPB-based reblurring loss:
\begin{equation}
	\label{Loss CDCN}
	L_{CDCN}=L_{content}+\lambda L_{fr}+L_{reblur}
\end{equation}
\begin{equation}
	\label{Loss ms-content}
	L_{content}=\sum_{i=1}^{3}\sum_{k=1}^{3} (\frac{1}{M}\sum_{m=1}^{M}\vert \vert (S_{i1k})^m-(S_{GT_{i1k}})^m\vert \vert_1)
\end{equation}
\begin{equation}
	\label{Loss ms-fr}
	L_{fr}=\sum_{i=1}^{3}\sum_{k=1}^{3}(\frac{1}{M}\sum_{m=1}^{M}\vert \vert F((S_{i1k})^m)-F((S_{GT_{i1k}})^m)\vert \vert_1)
\end{equation}
where, $(S_{i1k})^m$ and $(S_{GT_{i1k}})^m$ denote the $m-th$ recovered image and the $m-th$ ground truth sharp image at the $k-th$ DeBlock of the $1-th$ level of the $i-th$ scale, respectively. $\lambda$ is set to $0.1$ in our experiments. For the parameters of our proposed CDCN, in this paper, we propose a inter-scale parameter sharing scheme: the parameters between the levels, which have the same multi-patch model, are shared. And that, the parameters between the levels, which belong to the same scale are independent.

\subsection{The Differences to multi-input multi-output U-net (MIMO-UNet)}

The multi-input multi-output strategy and the fusion of different resolution features are also introduced in MIMO-UNet for SIDSBD. However, there are two main differences between MIMO-UNet and our CDCN. The first difference is the network architecture. In MIMO-UNet, Cho et al.\cite{cho2021rethinking} adopt a traditional single-scale coarse-to-fine strategy, where a input blurred image is encoded and decoded only once, and the output multiple deblurred images cannot be used to guide image restoration of other scales. While, our CDCN consists of three scales, where each scale contains two levels, therefore, a current-scale blurred image will be encoded and decoded twice. And, because of the multi-scale and multi-level architecture, in our CDCN, all intermediate deblurred results can be used to guide the image restoration at the next scale. The second difference is the estimation of the spatially-variant motion blur kernels. In MIMO-UNet, Cho et al.\cite{cho2021rethinking} directly remove the spatially-variant motion blurs from one inputted motion blurred image, without blur kernels estimation. Compared with MIMO-UNet, our CDCN can accurately estimate spatially-variant motion blur kernels from only one single motion blurred image without the optical flow. 

\subsection{The Differences to Spatially Variant Deconvolution Network (SVDN)}

Another related work to our CDCN is SVDN, where the deformable convolution is utilized to learn the sampling points of the blur kernels. Although our CDCN also adopts the deformable convolution, there are three main differences between SVDN and our CDCN. First, for the learning of the sampling points, SVDN uses the bi-directional optical flows to approximate the blur kernels, and makes the spatial distribution of the deformable sampling points close to the optical flows by using a distance-based loss function. Therefore, SVDN is still essentially a multi-frame-based blur kernel estimation method. On the contrary, our CDCN learns the sampling points of spatially-variant motion blur kernels from only one inputted motion blurred image by using a regular convolution without any optical flow. Therefore, our CDCN is essentially a single-image-based blur kernel estimation method, which is suitable for the SIDSBD well. Second, compared with SVDN, our CDCN proposes a PMPB-based reblurring loss function to constrain the accuracy of the learned sampling points, which makes the learned sampling points fit the trajectory of the relative motion of each pixel well and achieves accurate spatially-variant motion blur kernels estimation from a single motion blurred image. Third, for the deconvolution operation, SVDN uses two deformable convolutions to get the deblurred features. However, our CDCN can not only generate deblurred features by using only one deformable convolution, but also can achieve better deblurring performance for SIDSBD. 

\section{EXPERIMENTS}

To compare our method with the state-of-the-art SIDSBD methods, and demonstrate the effectiveness of our method, extensive experiments are performed on a PC with four NVIDIA Geforce RTX 3090 GPUs and the Intel Core I9-10980XE CPU, and the PyTorch 1.9.0 Library.

\subsection{The Datasets and Implementation Details}
In this paper, we train our model on the GOPRO\cite{nah2017deep} dataset and then test it on the GOPRO and HIDE\cite{HAdeblur} test datasets respectively. For the GoPro dataset, we use 2103 image pairs for training and the remaining 1111 pairs for testing, which is same as \cite{nah2017deep}. For the HIDE dataset, we use 2025 pairs for testing, which is same as \cite{HAdeblur}.

For training of our CDCN, the Adam\cite{kingma2014adam} with $\beta_1=0.9$,$\beta_2=0.999$ and $\varepsilon=1e-8$ is used as the optimizer to optimize our network for $2000$ epochs which are sufficient for convergence.  The learning rate is initially set to $1e-4$ and decreased by the factor of $0.5$ at every $200$ epochs. For every training iteration, we randomly sample eight images. $N$ and $M$ are set to $3$ and $7$ respectively. The number of residual blocks in each EnBlock and each DeBlock is 8. And unless otherwise specified, the entire network uses the $3\times 3$ convolution kernel for all other convolutional layers by default. For the quantitative metrics, we use the peak signal-to-noise ratio (PSNR) and the self-similarity measure (SSIM) to evaluate the performance of our method quantitatively.  For data augmentation, each patch was horizontally flipped with a probability of $0.5$.

\subsection{Ablation  Experiments}

\begin{table}[!t]
	\caption{ABLATION STUDY OF THE PROPOSED PMPB-BASED REBLURRING LOSS, THE CDCR MODULE AND THE MSML-MIMO ARCHITECTURE. THE PSNR AND SSIM ARE OBTAINED BY AVERAGING 1,111 GOPRO TESTING IMAGES\label{Ablation}}
	\centering
	\begin{tabular}{|c|c|c|}
		\hline
		Models        & PSNR  & SSIM  \\ \hline
		CDCN-NoPMPBReBlur & 32.08 & 0.952 \\ \hline
		CDCN-NoCDCR  & 31.87 & 0.948 \\ \hline
		CDCN-1level  & 31.98 & 0.951 \\ \hline
		CDCN-NoMIMO  & 32.18 & 0.954 \\ \hline
		CDCN          & 32.59 & 0.958  \\ \hline
	\end{tabular}
\end{table}

\begin{figure*}[!t]
	\centering
	\subfloat[with the PMPB-based reblurring constraint]{\includegraphics[width=2.2in]{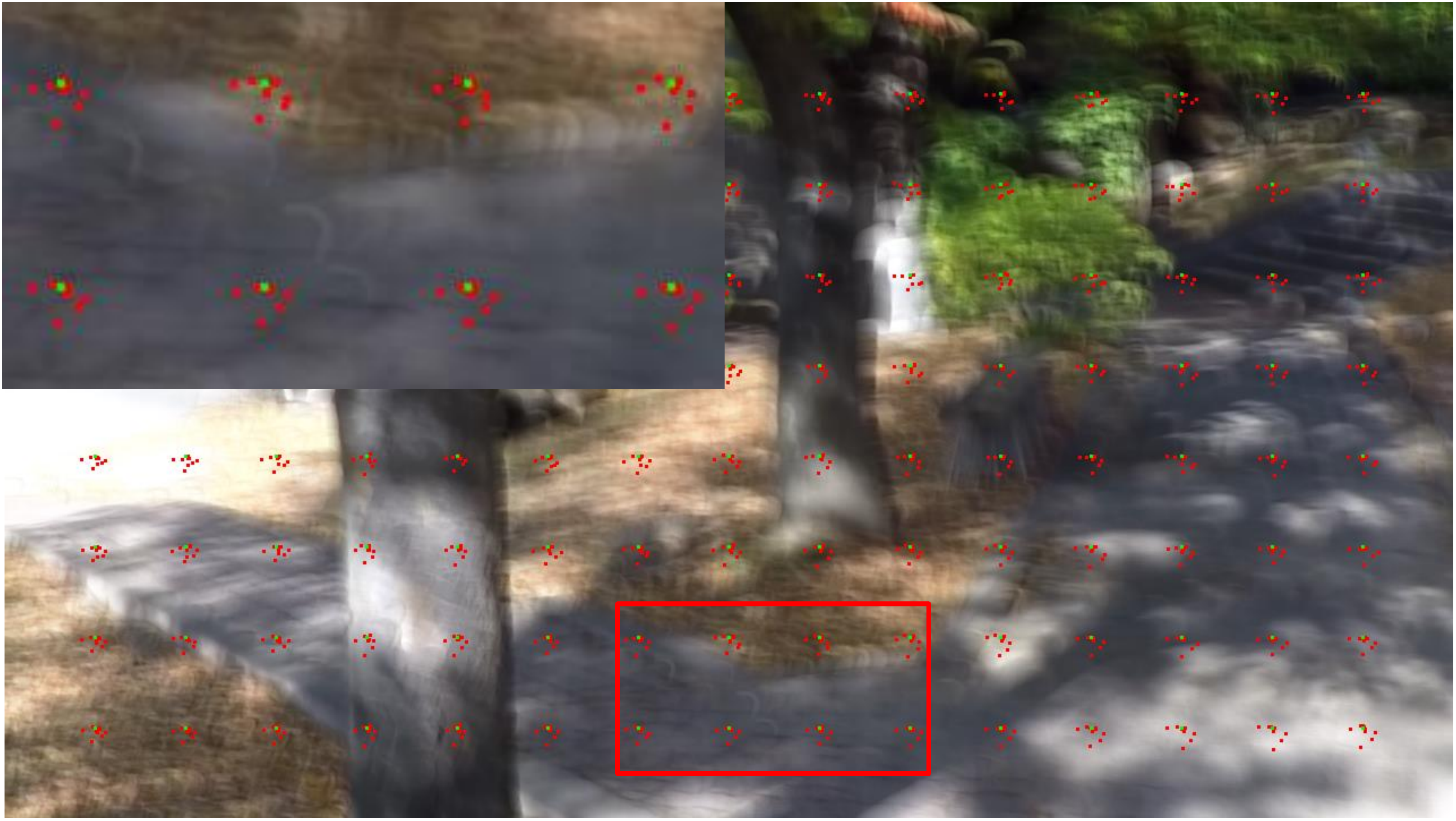}
		\label{BlurKernelReBlur}}
	\subfloat[with the PMPB-based reblurring constraint]{\includegraphics[width=2.2in]{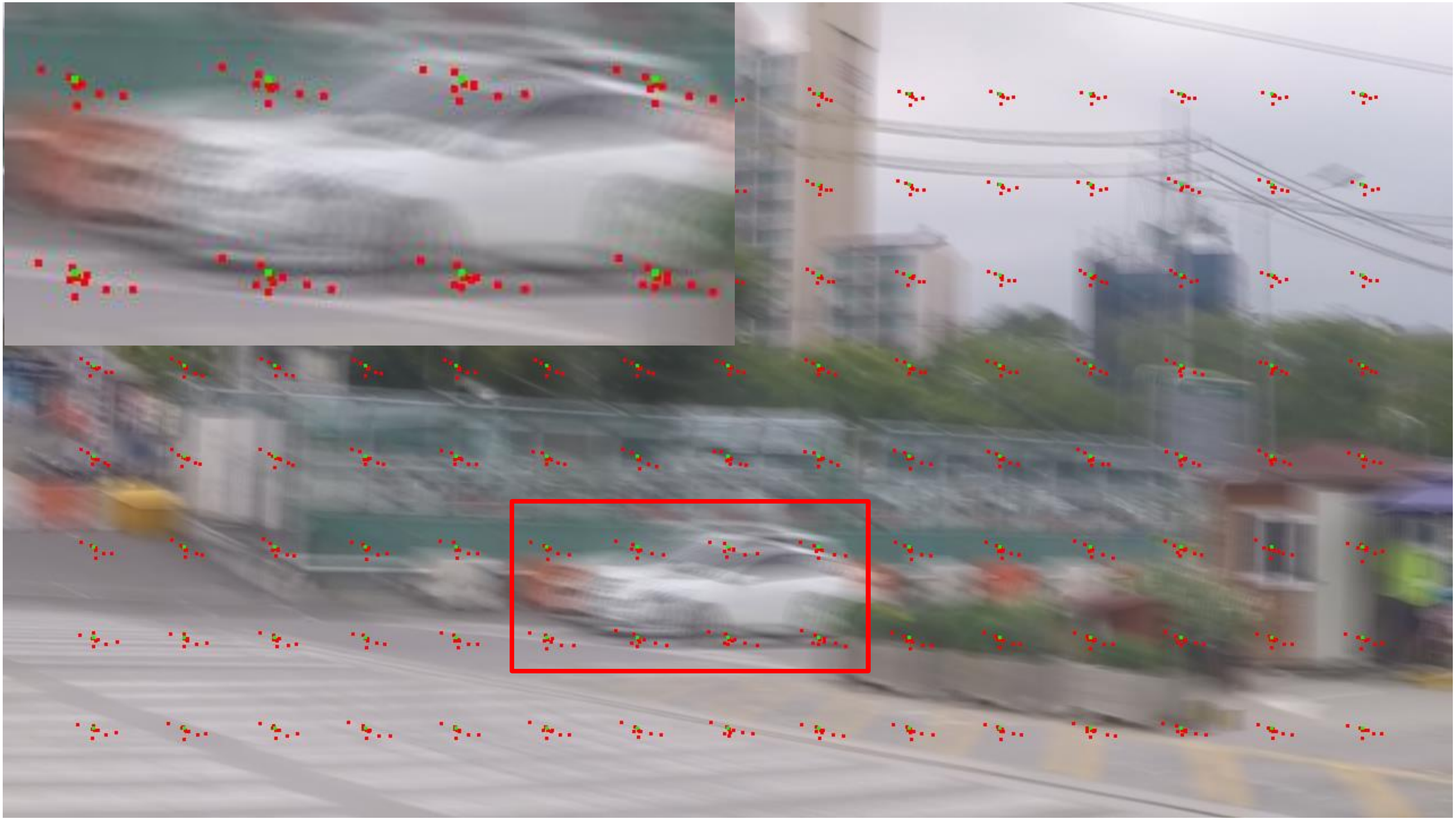}
		\label{BlurKernelReBlur2}}
	\subfloat[with the PMPB-based reblurring constraint]{\includegraphics[width=2.2in]{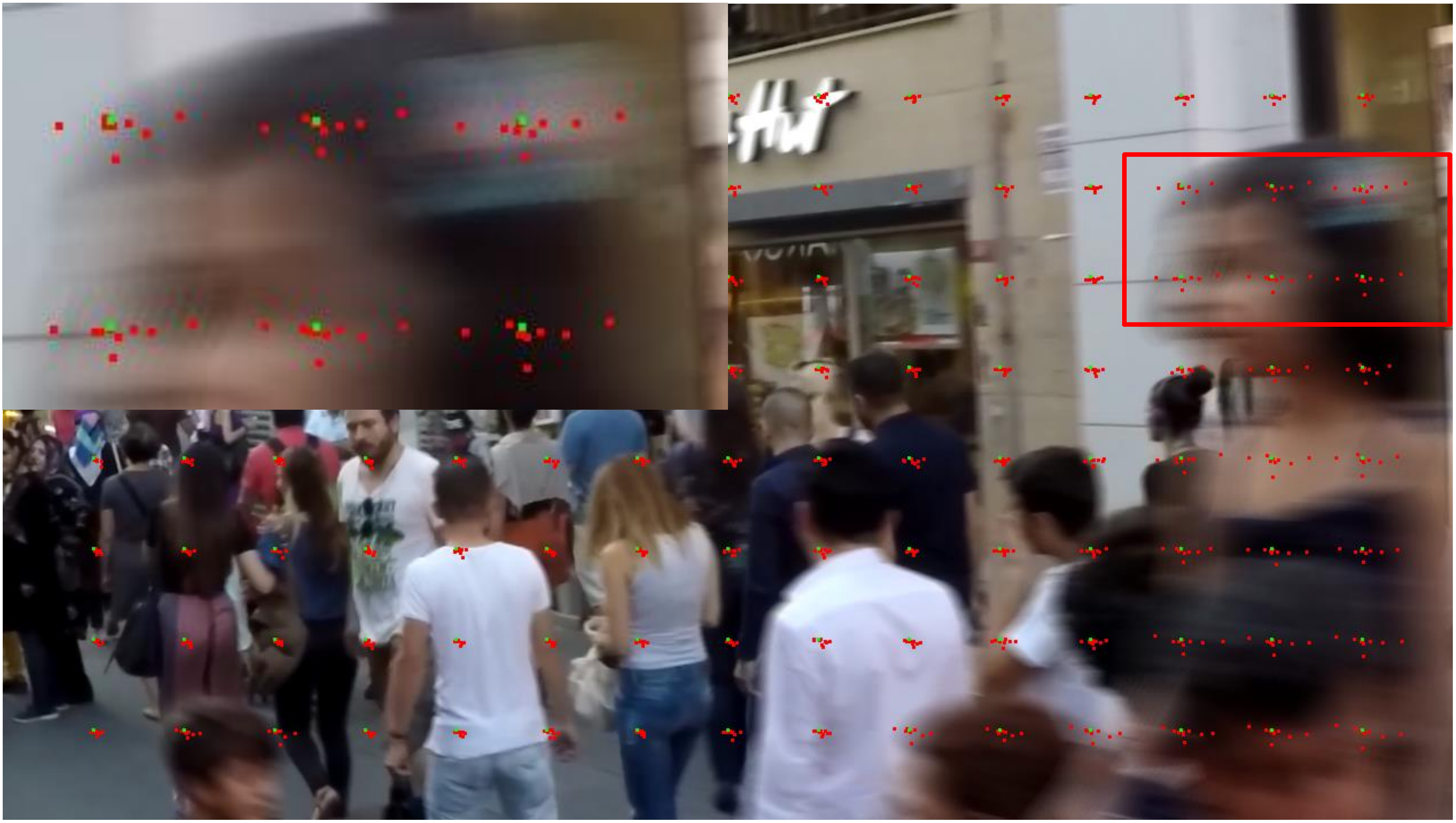}
		\label{BlurKernelReBlur3}}
	\hfil
	\subfloat[without the PMPB-based reblurring constraint]{\includegraphics[width=2.2in]{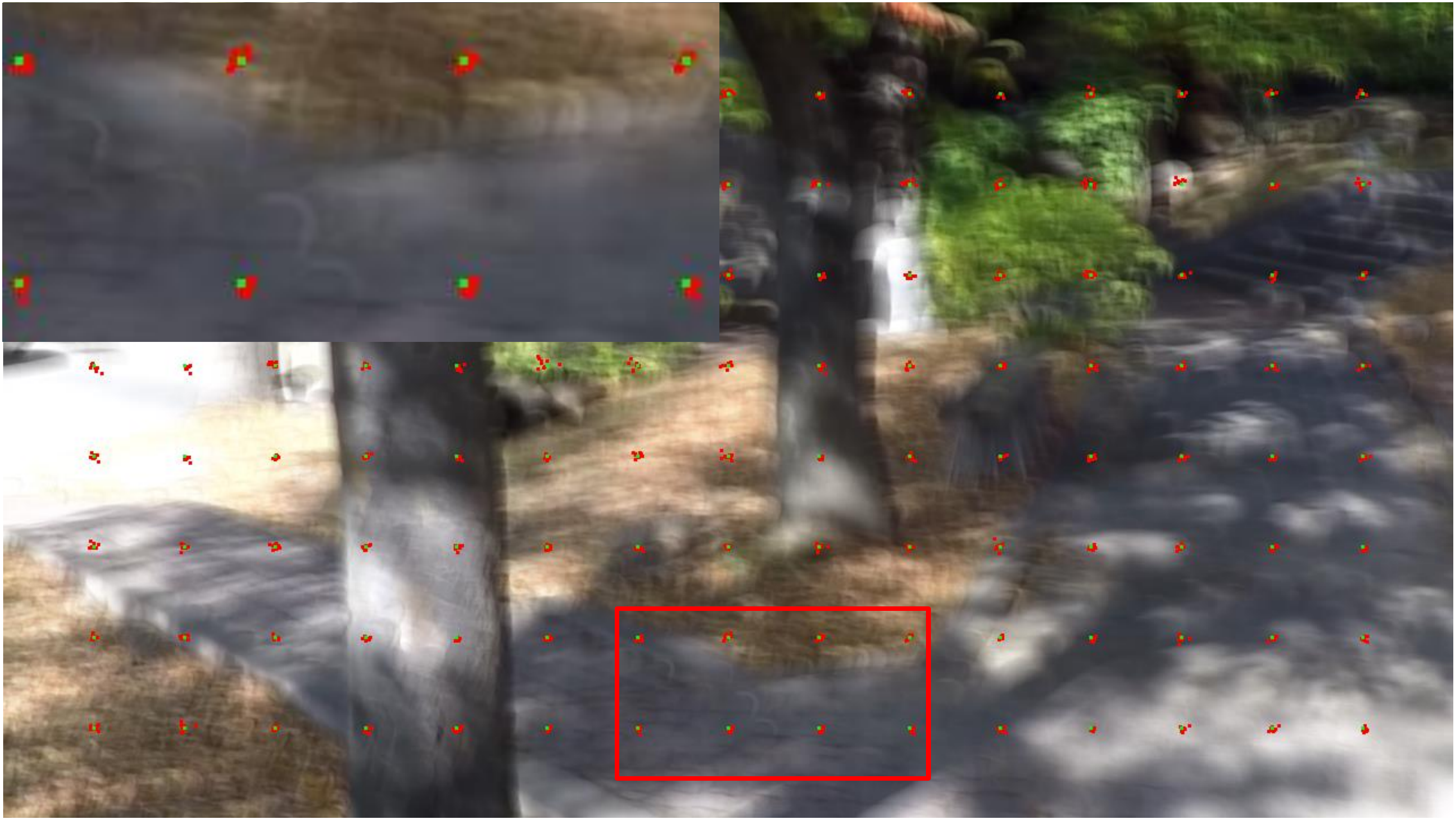}
		\label{BlurKernelNoReBlur}}
	\subfloat[without the PMPB-based reblurring constraint]{\includegraphics[width=2.2in]{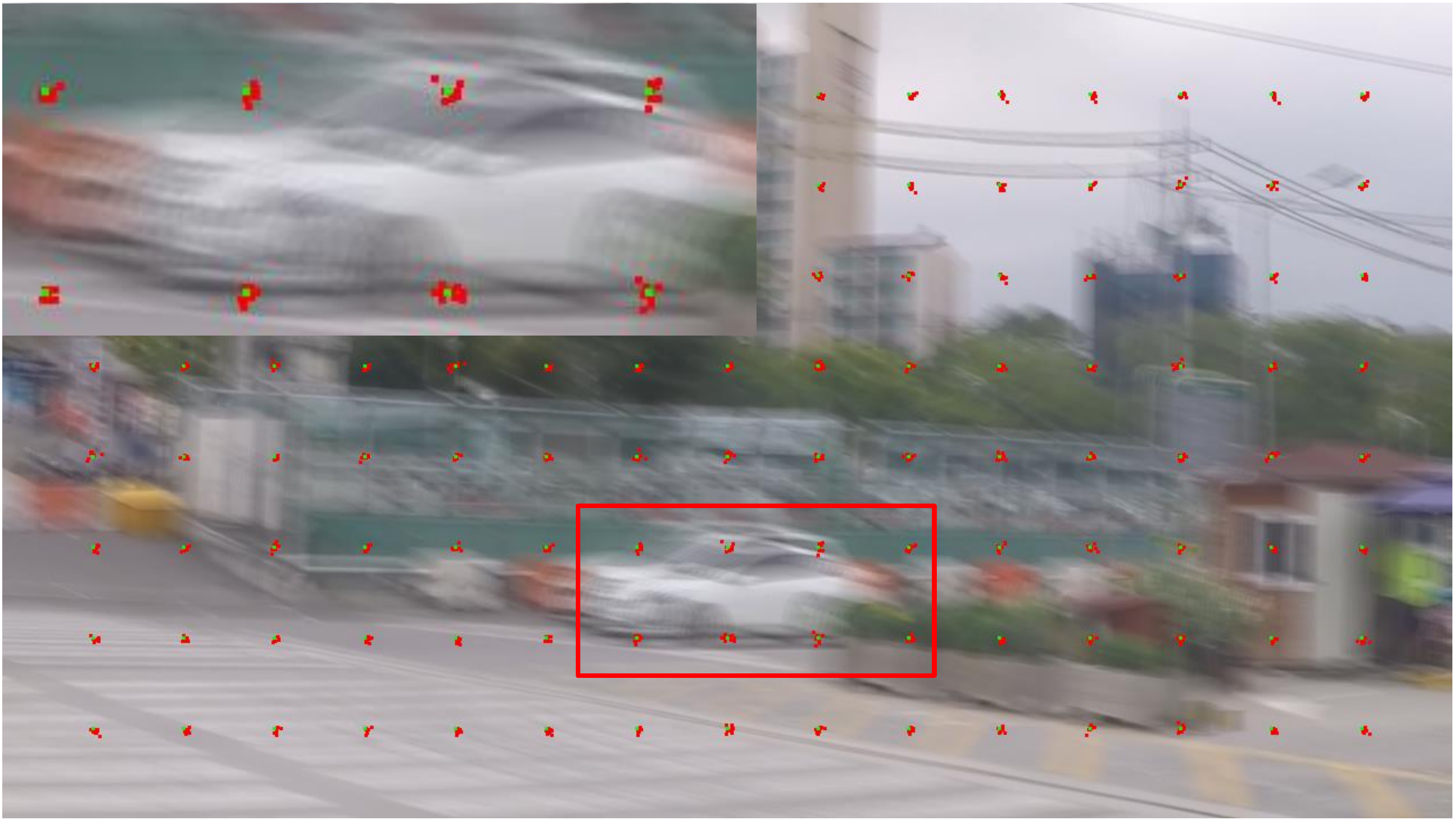}
		\label{BlurKernelNoReBlur2}}
	\subfloat[without the PMPB-based reblurring constraint]{\includegraphics[width=2.2in]{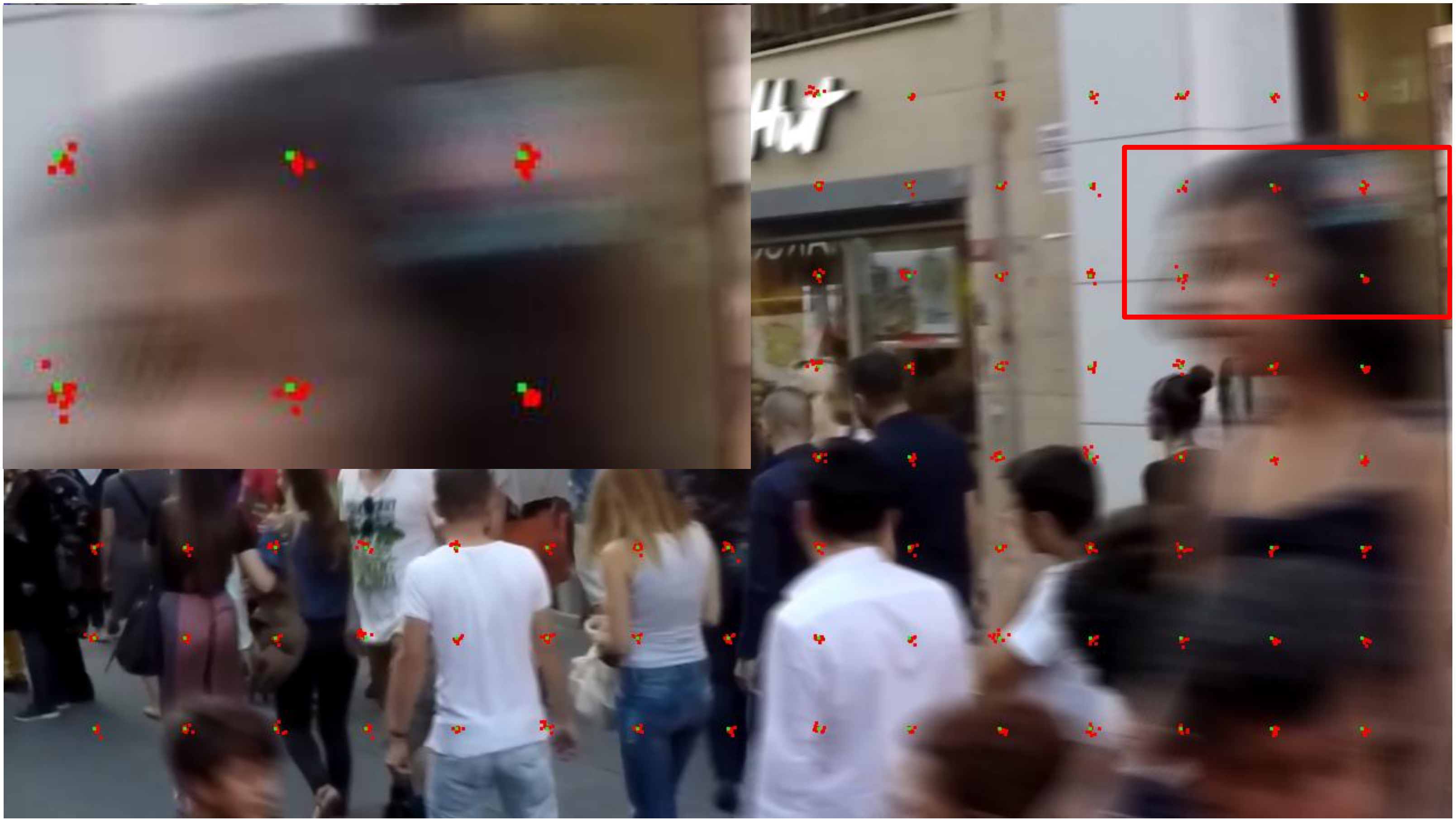}
		\label{BlurKernelNoReBlur3}}
	\hfil
	\caption{Visualization of the sampling points with and without the PMPB-based reblurring constraint.}
	\label{BlurKernel}
\end{figure*}

\begin{figure*}[!t]
	\centering
	\subfloat[CDCN-NoPMPBReBlur]{\includegraphics[width=2.2in]{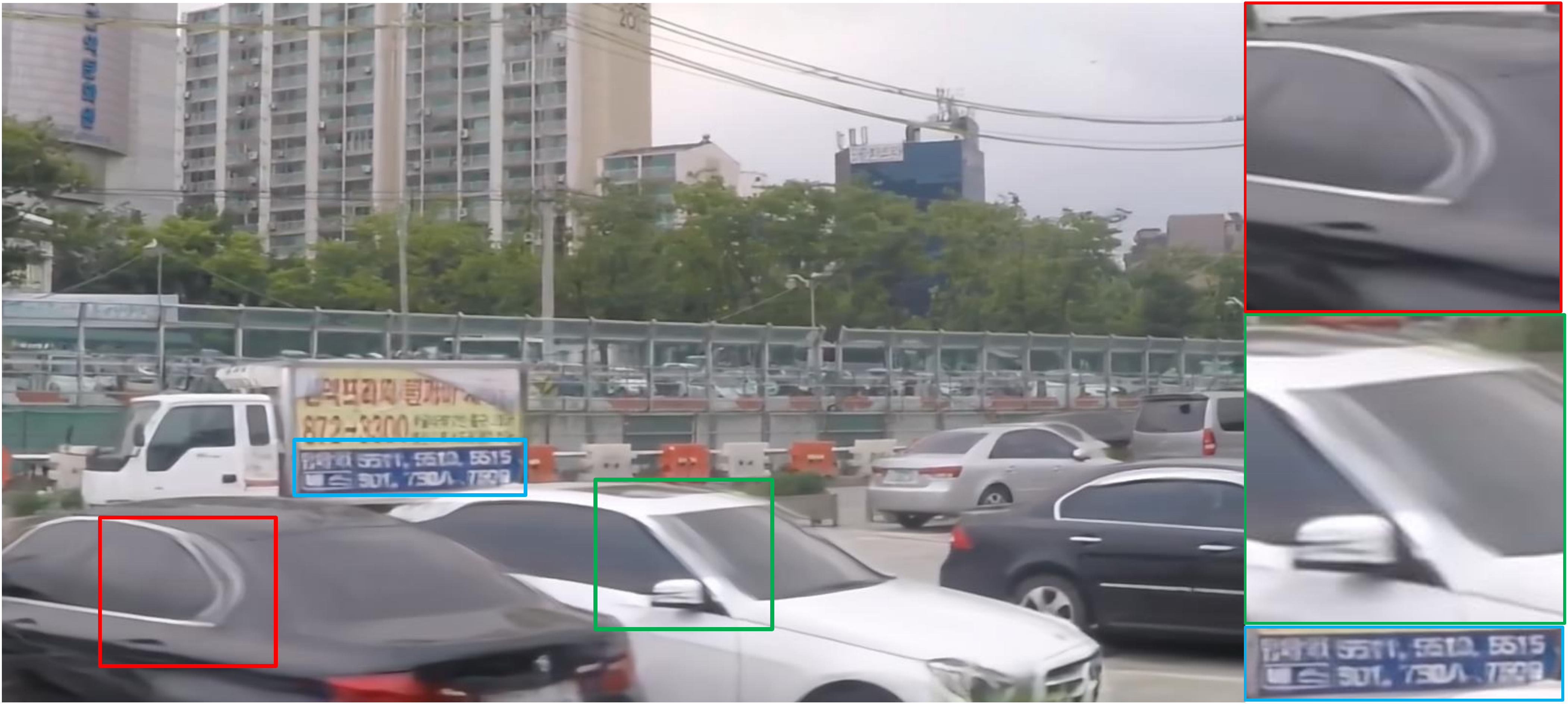}
		\label{CDCN-NoPMPBReBlur}}
	\subfloat[CDCN-NoCDCR]{\includegraphics[width=2.2in]{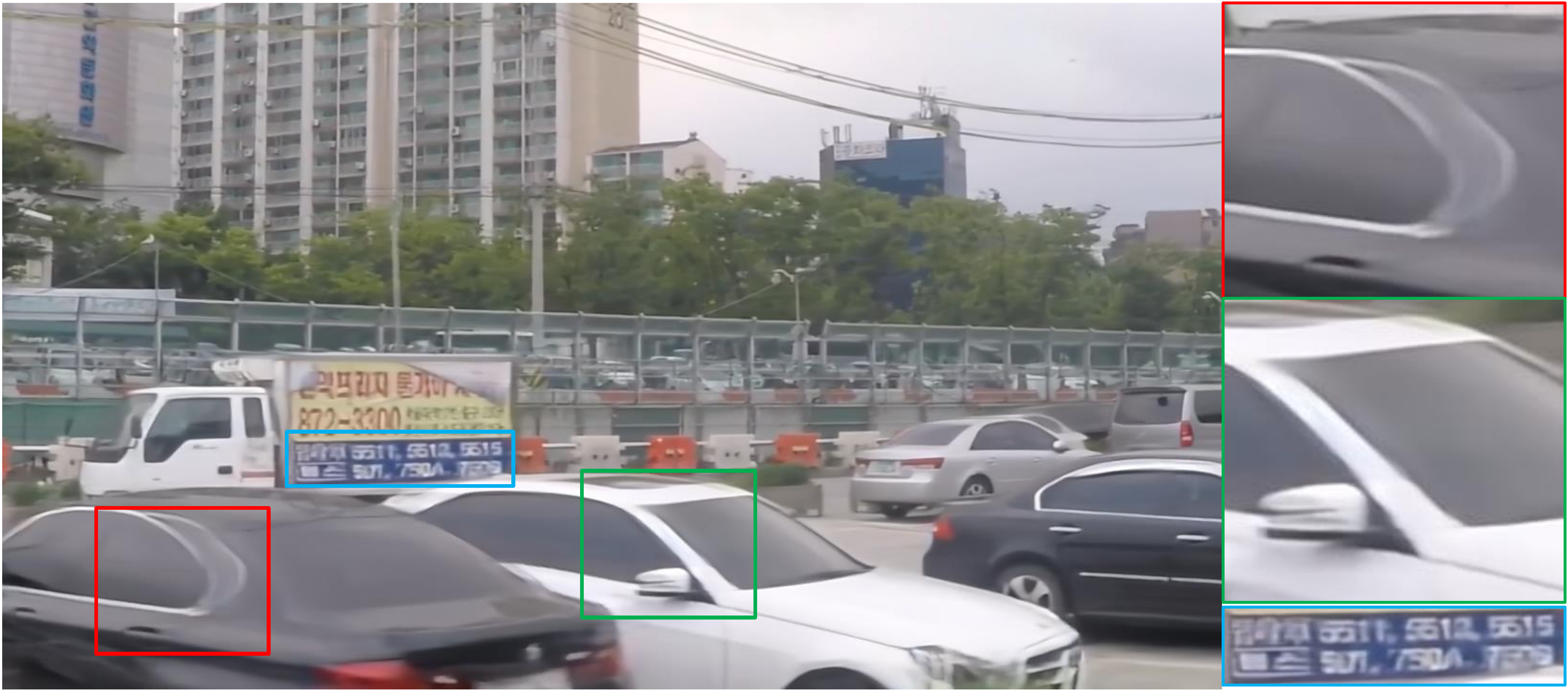}
		\label{CDCN-NoCDCR}}
	\subfloat[CDCN-1level]{\includegraphics[width=2.2in]{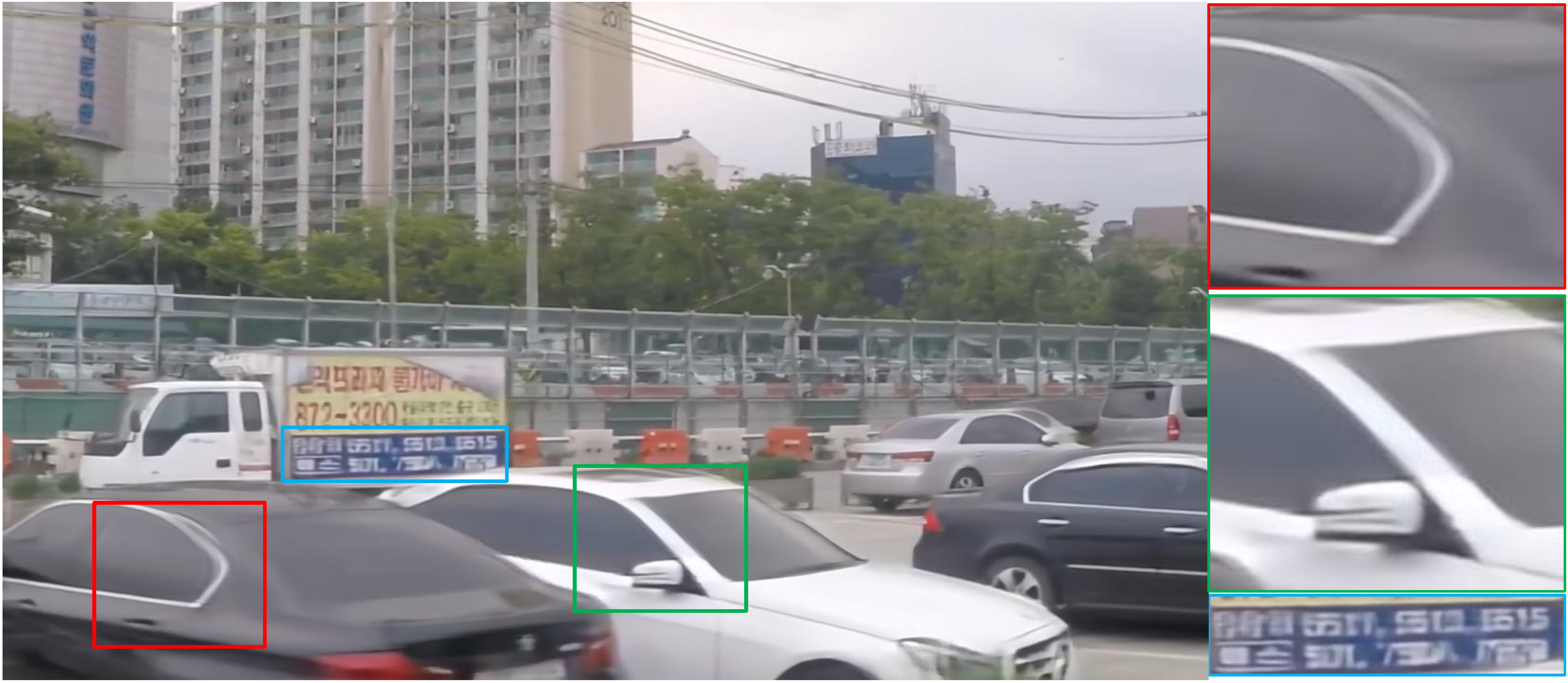}
		\label{CDCN-1level}}
	\hfil
	\subfloat[CDCN-NoMIMO]{\includegraphics[width=2.2in]{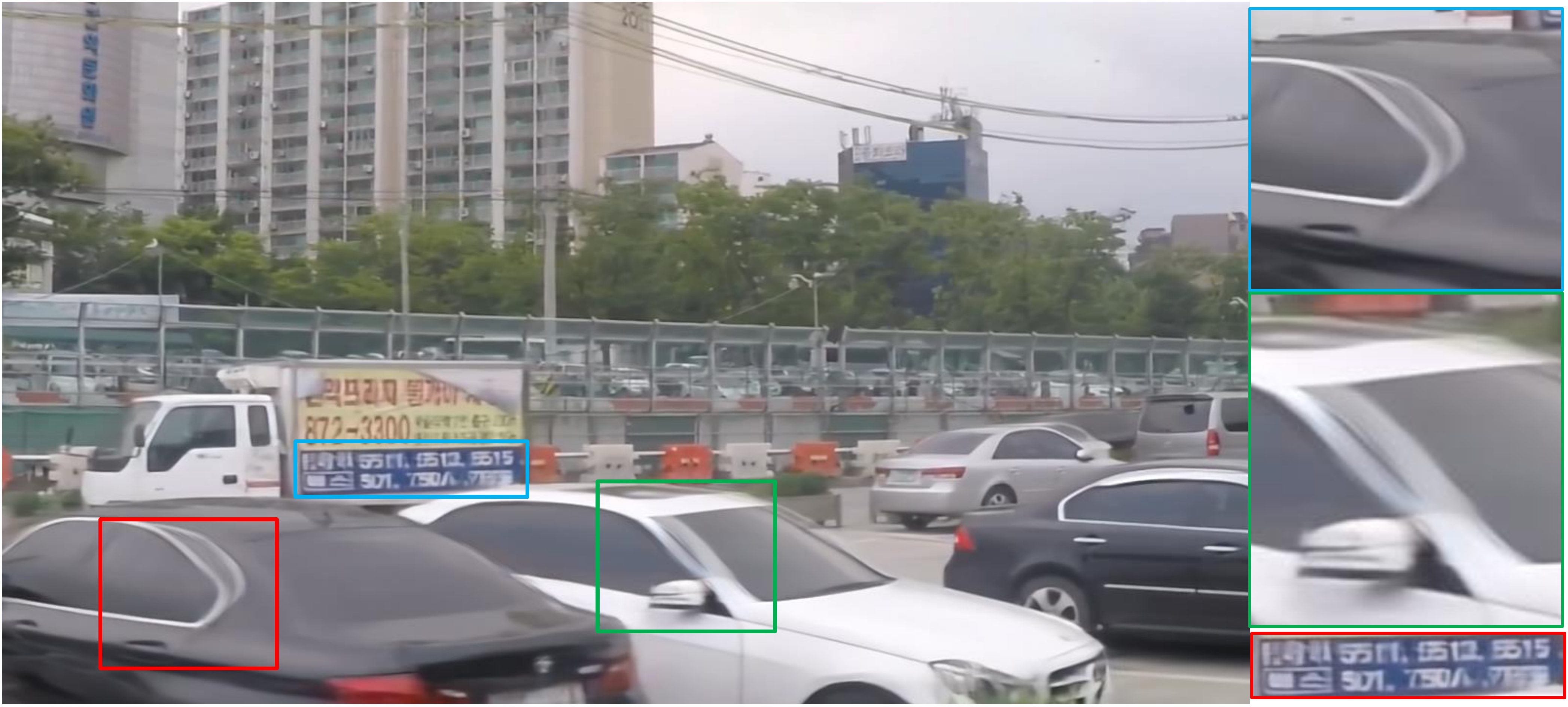}
		\label{CDCN-NoMIMO}}
	\subfloat[Our proposed CDCN]{\includegraphics[width=2.2in]{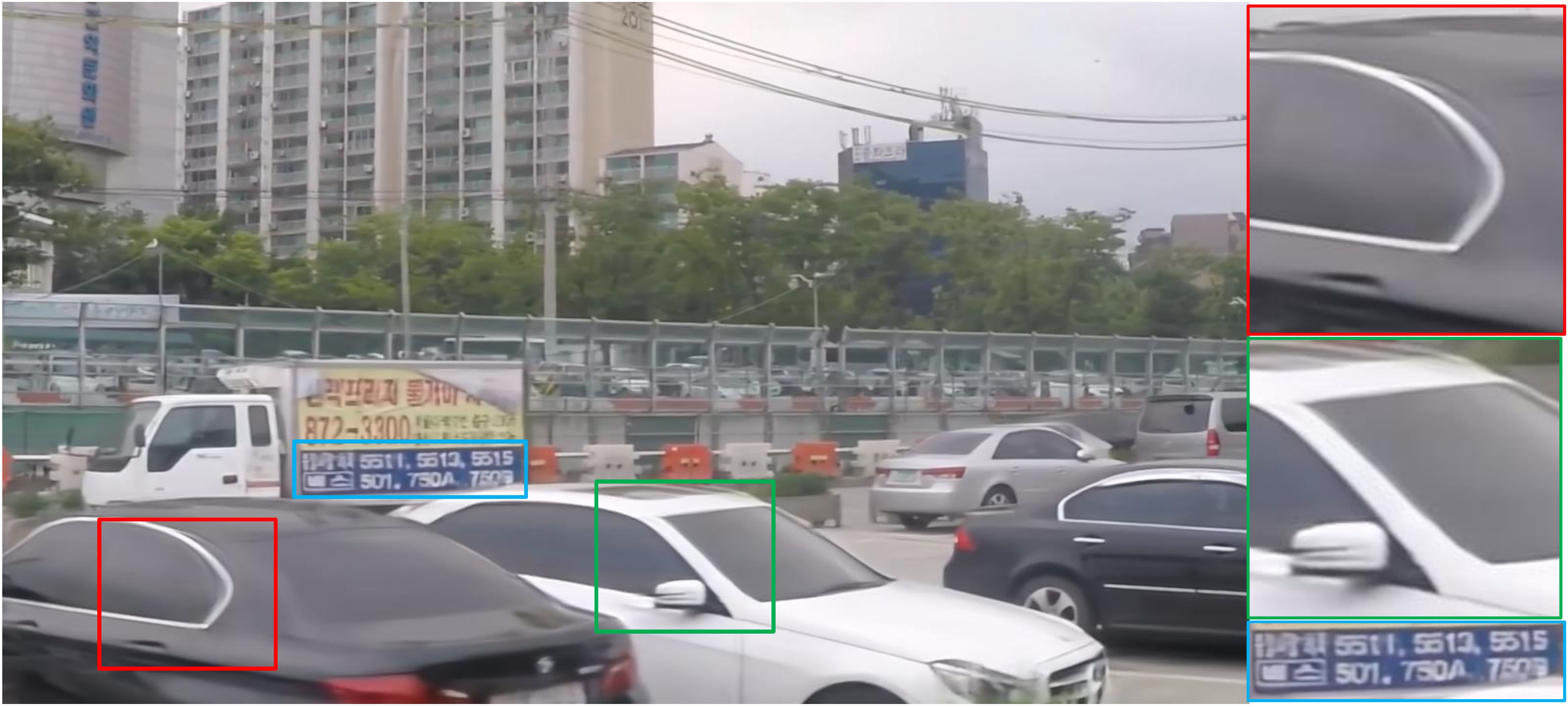}
		\label{CDCN-CDCN}}
	\hfil
	\caption{The subjective visual evaluation results of the models CDCN-NoPMPBReBlur, CDCN-NoCDCR, CDCN-1level, CDCN-NoMIMO and CDCN.   }
	\label{AblationVisual}
\end{figure*}

As we discussed above, the main contributions of our proposed CDCN are: the PMPB-based reblurring loss function for constraining the convergence of the sampling points of spatially-variant motion blur kernels, a CDCR module for predicting the inverse kernels and generating deblurred features, and an MSML-MIMO architecture for more powerful features extraction ability. So, in this subsection, we will individually test these three parts to better understand the effectiveness of each part via the experiments.

1) Evaluation of the PMPB-based reblurring loss: As shown in Equations (\ref{ReBlur Loss}), the proposed PMPB-based reblurring loss function can constrain the solution spaces of the $offset_{bk}$ and $weight_{bk}$, and make the learned sampling points fit the trajectory of the relative motion of each pixel well. In order to verify the effectiveness of the PMPB-based reblurring loss, we compare our proposed CDCN with a model named CDCN-NoPMPBReBlur, which removes Equations (\ref{ReBlur Loss}) from the proposed CDCN (i.e. without any constraint on the $offset_{bk}$ and $weight_{bk}$). In other words, the values of $offset_{bk}$ and $weight_{bk}$ will be completely self-learned. The CDCN-NoPMPBReBlur model shares the same framework as Fig. \ref{CDCR} and is trained on the same GoPro training dataset as the proposed CDCN model. Table \ref{Ablation} shows that without the PMPB-based reblurring constraint, the PSNR and SSIM of the CDCN-NoPMPBReBlur model reduce by 0.51dB and 0.006, respectively. 

Fig. \ref{BlurKernel}  shows the comparisons of the distribution of sampling points with and without the PMPB-based reblurring constraint. On the one hand, from Fig. \ref{BlurKernelReBlur2}, we can see that, the light and shadow on the ground provides clear information of the relative movement trajectory, and our proposed PMPB-based reblurring loss can constrain the learned sampling points fit the trajectory of the relative motion of each pixel very well: the shape of the red points is very similar to that of the light and shadow (Please see the Fig. \ref{BlurKernelReBlur}, \ref{BlurKernelReBlur2} and \ref{BlurKernelReBlur3}  and the zoomed in regions). On the other hand, from Fig. \ref{BlurKernelReBlur3}  we can see that, our proposed PMPB-based reblurring loss can also make the sampling points change with the level of the blur: different blurred regions have different sampling point distributions, which are consist with the degrees of the blurs (Please see the left region, the middle region and the right region of Fig. \ref{BlurKernelReBlur3} and the zoomed in region). However, the CDCN-NoPMPBReBlur model can not learn any information about the spatially-variant motion blur kernel (Please see the Figs. \ref{BlurKernelNoReBlur}, \ref{BlurKernelNoReBlur2} and \ref{BlurKernelNoReBlur3}, and the zoomed in regions)

2) Evaluation of the CDCR module: As shown in Fig. \ref{CDCR}, we use two PReLU layers and two regular convolutional layers to predict the inverse kernels from the learned $offset_{bk}$ and $weight_{bk}$, and then the predicted inverse kernels are used to generate the deblurred features via a deformable convolution. To demonstrate the effectiveness of the CDCR module, we compare our proposed CDCN with a model named CDCN-NoCDCR, which removes $Conv_2()$ , $Conv_3()$ , two PReLU layers, and the final deformable convolution from the CDCR module, and only keep the $Conv_1()$ and the $SoftMax$ for calculating PMPB-based reblurring loss. Therefore, in the CDCN-NoCDCR model, $DB_{i21}^{in}=EB_{i23}^{out}$ and $DB_{i11}^{in}=EB_{i23}^{out}\oplus EB_{i13}^{out}$. The CDCN-NoCDCR model is trained on the same GoPro training dataset as the proposed CDCN model. As can be seen from Table \ref{Ablation}, without the prediction of the inverse kernels, the PSNR and SSIM of the CDCN-NoCDCR model are reduced by 0.72dB and 0.01, respectively.

3) Evaluation of the MSML-MIMO architecture: To demonstrate the advantage of our MSML-MIMO architecture, we compare our proposed CSCN with $2$ baseline models: the CDCN-1level model, where each scale only contains one level, and the CDCN-NoMIMO, where only $B_{i11}$ and $B_{i21}$ are considered as the inputs of the $i-th$ scale, and only $S_{i13}$ is considered as the output of the $i-th$ scale. The models CDCN-1level and CDCN-NoMIMO are trained on the same GoPro training dataset as the proposed CDCN model respectively. From Table \ref{Ablation} we can see that, on the one hand, compared with our CDCN model, CDCN-1level model can not conduct the residual operation between levels, and the values of PSNR and SSIM are reduced by 0.61dB and 0.007, respectively. On the other hand, without more input information and intermediate restoration results, the PSNR and SSIM of the CDCN-NoMIMO model are reduced by 0.41dB and 0.004, respectively.

Fig. \ref{AblationVisual} illustrates the subjective visual evaluation results of the models CDCN-NoPMPBReBlur, CDCN-NoCDCR, CDCN-1level, CDCN-NoMIMO and our proposed CDCN. We could found that our proposed CDCN can obtain higher quality restoration image, which has sharper and clearer edges. Table \ref{Ablation} and Fig. \ref{AblationVisual} demonstrate that all contributions play important roles in estimating accurate blur kernels and recovering of high quality images.

\subsection{The Comparisons With the State-of-the-Art SIDSBD Methods on the Synthetic Benchmark Datasets}

\begin{table}[!t]
	\caption{ALL THE MODELS. ALL MODELS ARE TRAINED ONLY ON THE GOPRO\cite{nah2017deep} TRAINING IMAGE PAIRS AND DIRECTLY APPLIED TO THE HIDE\cite{HAdeblur} TESTING DATASETS.\label{GOPRO_HIDE}}
	\centering
	\begin{tabular}{|c|cc|cc|}
		\hline
		\multirow{2}{*}{Method} & \multicolumn{2}{c|}{GOPRO}                      & \multicolumn{2}{c|}{HIDE}                  \\ \cline{2-5} 
		& \multicolumn{1}{c|}{PSNR}           & SSIM      & \multicolumn{1}{c|}{PSNR}      & SSIM      \\ \hline
		Xu et al.\cite{xu2013unnatural} & \multicolumn{1}{c|}{21.00}  & 0.741     & \multicolumn{1}{c|}{-}         & -         \\ \hline
		DeblurGAN\cite{kupyn2018deblurgan} & \multicolumn{1}{c|}{28.70} & 0.858   & \multicolumn{1}{c|}{24.51}     & 0.871     \\ \hline
		Nah et al.\cite{nah2017deep} & \multicolumn{1}{c|}{29.08}     & 0.914     & \multicolumn{1}{c|}{25.73}     & 0.874     \\ \hline
		Zhang et al.\cite{zhang2018dynamic} & \multicolumn{1}{c|}{29.19} & 0.931  & \multicolumn{1}{c|}{-}         & -         \\ \hline
		DeblurGAN-v2\cite{kupyn2019deblurgan} & \multicolumn{1}{c|}{29.55} & 0.934 & \multicolumn{1}{c|}{26.61}    & 0.875     \\ \hline
		SRN\cite{tao2018scale}  & \multicolumn{1}{c|}{30.26}          & 0.934     & \multicolumn{1}{c|}{28.36}     & 0.915     \\ \hline
		Gao et al.\cite{gao2019dynamic} & \multicolumn{1}{c|}{30.90}  & 0.935     & \multicolumn{1}{c|}{29.11}     & 0.913     \\ \hline
		DBGAN\cite{zhang2020deblurring} & \multicolumn{1}{c|}{31.10}  & 0.942     & \multicolumn{1}{c|}{28.94}     & 0.915     \\ \hline
		MT-RNN\cite{park2020multi} & \multicolumn{1}{c|}{31.15}       & 0.945     & \multicolumn{1}{c|}{29.15}     & 0.918     \\ \hline
		DMPHN\cite{zhang2019deep} & \multicolumn{1}{c|}{31.20}        & 0.940     & \multicolumn{1}{c|}{29.09}     & 0.924     \\ \hline
		MSCAN\cite{wan2020deep} & \multicolumn{1}{c|}{31.24}        & 0.945     & \multicolumn{1}{c|}{29.63}     &  0.921    \\ \hline
		Suin et al.\cite{suin2020spatially} & \multicolumn{1}{c|}{31.85} & 0.948  & \multicolumn{1}{c|}{29.98}     & 0.930     \\ \hline
		SPAIR\cite{purohit2021spatially} & \multicolumn{1}{c|}{32.06} & 0.953    & \multicolumn{1}{c|}{30.29}     & 0.931     \\ \hline
		MIMO-UNet+\cite{cho2021rethinking} & \multicolumn{1}{c|}{32.45} & 0.957    & \multicolumn{1}{c|}{29.99}     & 0.930     \\ \hline
		\textbf{CDCN} & \multicolumn{1}{c|}{\textbf{32.59}} & \textbf{0.958} & \multicolumn{1}{c|}{\textbf{30.55}} & \textbf{0.935} \\ \hline
	\end{tabular}
\end{table}

\begin{figure*}[!t]
	\centering
	\subfloat[Blurry image]{\includegraphics[width=1.25in]{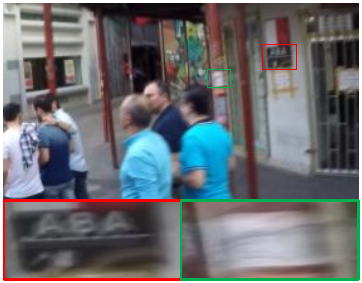}
		\label{Blur1}}
	\subfloat[Xu et al.\cite{xu2013unnatural}]{\includegraphics[width=1.25in]{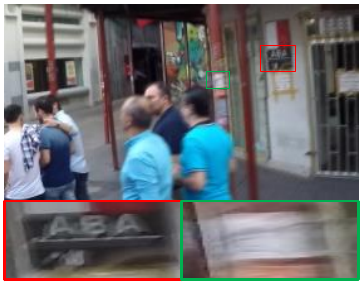}
		\label{Xu1}}
	\subfloat[Nah et al.\cite{nah2017deep}]{\includegraphics[width=1.25in]{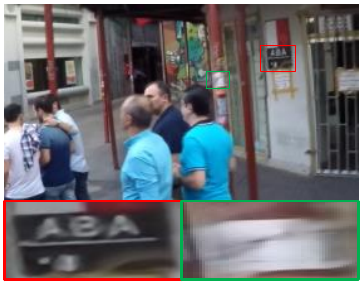}
		\label{Nah1}}
	\subfloat[DeblurGAN\cite{kupyn2018deblurgan}]{\includegraphics[width=1.25in]{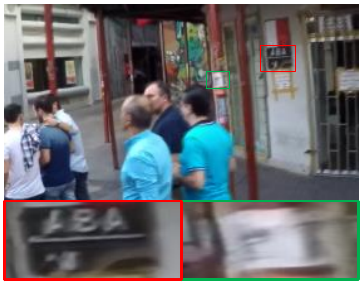}
		\label{DeblurGAN1}}
	\subfloat[SRN\cite{tao2018scale}]{\includegraphics[width=1.25in]{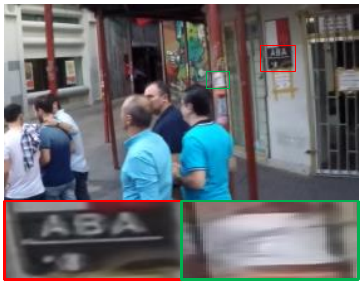}
		\label{SRN1}}
	\hfil
	\subfloat[Gao et al.\cite{gao2019dynamic}]{\includegraphics[width=1.25in]{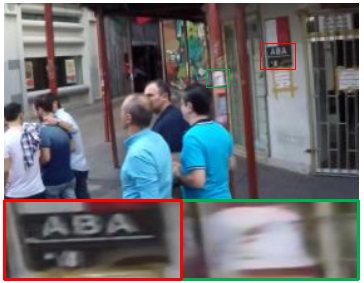}
		\label{Gao1}}
	\subfloat[Cai et al.\cite{cai2020dark}]{\includegraphics[width=1.25in]{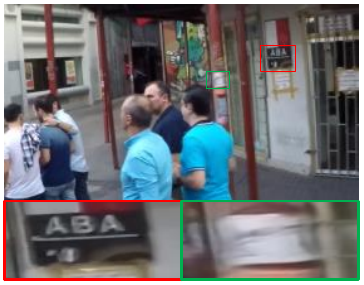}
		\label{Cai1}}
	\subfloat[MSCAN\cite{wan2020deep}]{\includegraphics[width=1.25in]{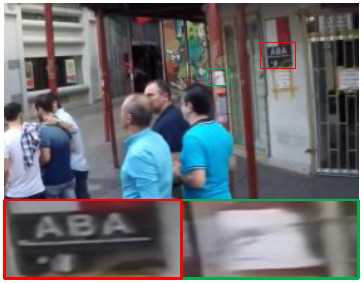}
		\label{MSCAN1}}
	\subfloat[MIMO-UNet+\cite{cho2021rethinking}]{\includegraphics[width=1.25in]{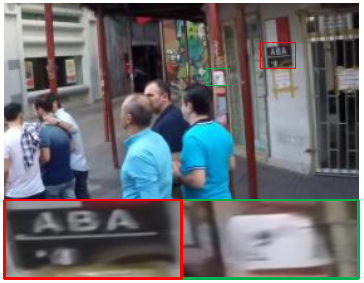}
		\label{MIMO-UNet+1}}
	\subfloat[Our CDCN]{\includegraphics[width=1.25in]{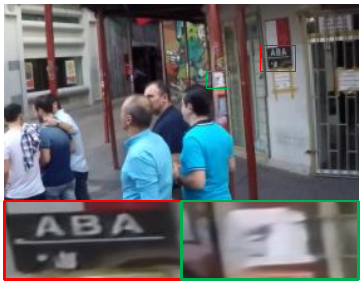}
		\label{CDCN1}}
	\caption{The qualitative evaluation comparisons of all the methods on the GoPro testing dataset.}
	\label{GOPRO comparisons1}
\end{figure*}

\begin{figure*}[!t]
	\centering
	\subfloat[Blur image]{\includegraphics[width=1.25in]{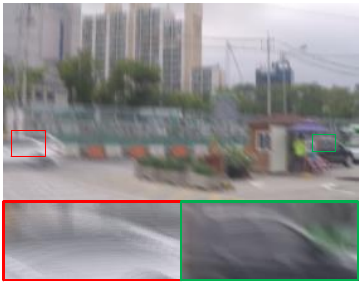}
		\label{Blur2}}
	\subfloat[Xu et al.\cite{xu2013unnatural}]{\includegraphics[width=1.25in]{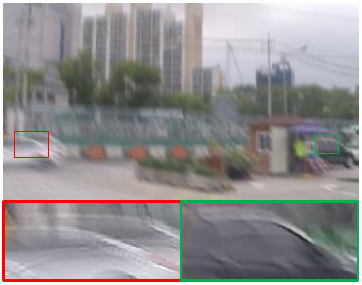}
		\label{Xu2}}
	\subfloat[Nah et al.\cite{nah2017deep}]{\includegraphics[width=1.25in]{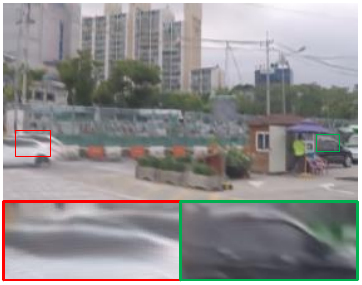}
		\label{Nah2}}
	\subfloat[DeblurGAN\cite{kupyn2018deblurgan}]{\includegraphics[width=1.25in]{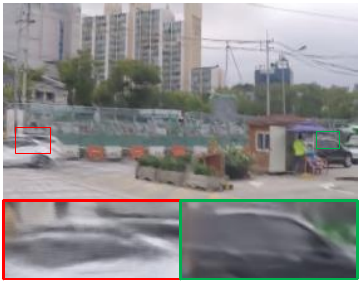}
		\label{DeblurGAN2}}
	\subfloat[SRN\cite{tao2018scale}]{\includegraphics[width=1.25in]{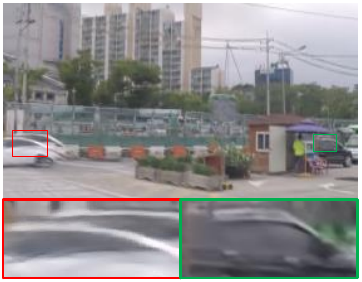}
		\label{SRN2}}
	\hfil
	\subfloat[Gao et al.\cite{gao2019dynamic}]{\includegraphics[width=1.25in]{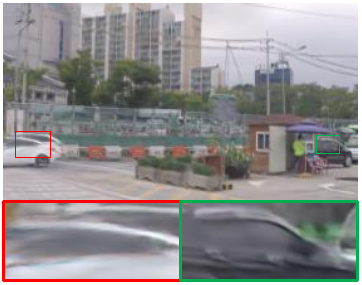}
		\label{Gao2}}
	\subfloat[Cai et al.\cite{cai2020dark}]{\includegraphics[width=1.25in]{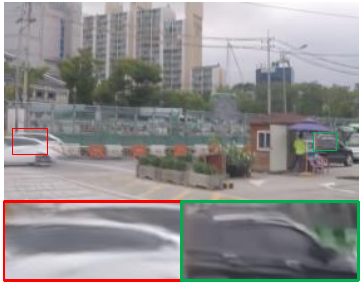}
		\label{Cai2}}
	\subfloat[MSCAN\cite{wan2020deep}]{\includegraphics[width=1.25in]{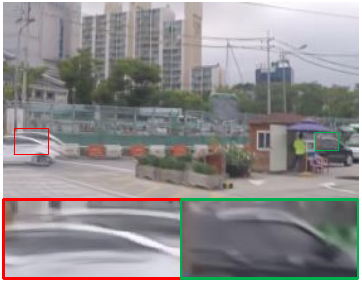}
		\label{MSCAN2}}
	\subfloat[MIMO-UNet+\cite{cho2021rethinking}]{\includegraphics[width=1.25in]{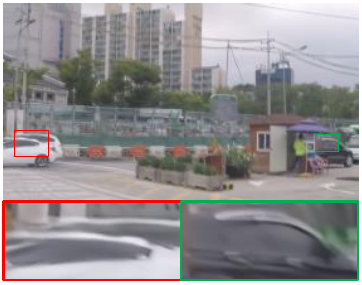}
		\label{MIMO-UNet+2}}
	\subfloat[Our CDCN]{\includegraphics[width=1.25in]{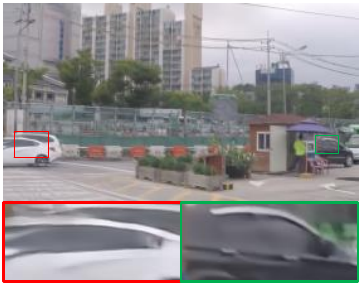}
		\label{CDCN2}}
	\caption{The qualitative evaluation comparisons of all the methods on the GoPro testing dataset.}
	\label{GOPRO comparisons2}
\end{figure*}

For the synthetic benchmark datasets comparisons, we use two synthetic datasets: the GoPro\cite{nah2017deep} and the HIDE\cite{HAdeblur}, and compare our method with fourteen state-of-the-art SIDSBD methods (Xu et al.\cite{xu2013unnatural}, DeblurGAN\cite{kupyn2018deblurgan}, Nah et al.\cite{nah2017deep}, Zhang et al.\cite{zhang2018dynamic}, DeblurGAN-v2\cite{kupyn2019deblurgan}, SRN\cite{tao2018scale}, Gao et al.\cite{gao2019dynamic}, DBGAN\cite{zhang2020deblurring}, MT-RNN\cite{park2020multi}, DMPHN\cite{zhang2019deep}, MSCAN\cite{wan2020deep}, Suin et al.\cite{suin2020spatially}, SPAIR\cite{purohit2021spatially} , MIMO-UNet+\cite{cho2021rethinking}). For fair comparison, the models \cite{kupyn2018deblurgan}, \cite{nah2017deep}, \cite{zhang2018dynamic}, \cite{kupyn2019deblurgan}, \cite{tao2018scale}, \cite{gao2019dynamic}, \cite{zhang2020deblurring}, \cite{park2020multi}, \cite{zhang2019deep}, \cite{wan2020deep},\cite{suin2020spatially}, \cite{purohit2021spatially} and \cite{cho2021rethinking} are trained on the 2103 GoPro training image pairs. Because the method \cite{xu2013unnatural} is an optimization-based method, we use the executable program provided by Xu et al. for the comparison. Table \ref{GOPRO_HIDE} shows the values of the mean PSNR and the mean SSIM of all models on the 1111 GoPro testing image pairs and 2025 HIDE testing image pairs. From Table \ref{GOPRO_HIDE} we can see that our CDCN significantly outperforms all the methods and could attain the highest mean PSNR value and mean SSIM value on both the GoPro testing images and the HIDE testing images. So, in summary, our CDCN can produce better deblurring results than the state-of-the-art SIDSBD methods in terms of the quantitative metrics.

Because of the space limitation, here, we only use two GoPro testing images (Fig. \ref{GOPRO comparisons1} and Fig. \ref{GOPRO comparisons2}) to illustrate the qualitative evaluation comparisons of the models \cite{xu2013unnatural} , \cite{kupyn2018deblurgan}, \cite{nah2017deep}, \cite{tao2018scale}, \cite{gao2019dynamic}, \cite{cai2020dark}, \cite{wan2020deep}, \cite{cho2021rethinking} and our CDCN model. From Fig. \ref{GOPRO comparisons1} and Fig. \ref{GOPRO comparisons2} we can see that, the deblurred images by models \cite{xu2013unnatural} , \cite{kupyn2018deblurgan}, \cite{nah2017deep}, \cite{tao2018scale}, \cite{gao2019dynamic}, \cite{cai2020dark}, \cite{wan2020deep}, \cite{cho2021rethinking} suffer from one or more of different degrees flaws: the blur, the distortion and the deformation, respectively. By contrast, our CDCN can obtain the highest quality restoration images: can not only remove various artifacts effectively, but also can recover sharper edges and more details. Please see the Figs. \ref{Xu1}-\ref{MIMO-UNet+1}, and the Figs. \ref{Xu2}-\ref{MIMO-UNet+2}, and the corresponding zoomed in regions. Table \ref{GOPRO_HIDE}, Fig. \ref{GOPRO comparisons1} and Fig. \ref{GOPRO comparisons2} demonstrate the superiority of our method on the synthetic benchmark datasets. More experimental results for synthetic benchmark datasets can be available at https://github.com/wuyang1002431655/CDCN.

\subsection{ The Comparisons with the state-of-the-art methods on the real blurred images}

\begin{figure*}[!t]
	\centering
	\subfloat[Blur image]{\includegraphics[width=1.25in]{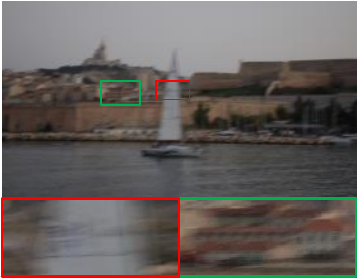}
		\label{Blur7}}
	\subfloat[SRN\cite{tao2018scale}]{\includegraphics[width=1.25in]{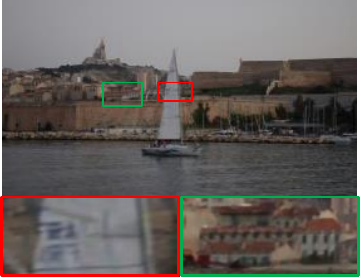}
		\label{SRN7}}
	\subfloat[Gao et al.\cite{gao2019dynamic}]{\includegraphics[width=1.25in]{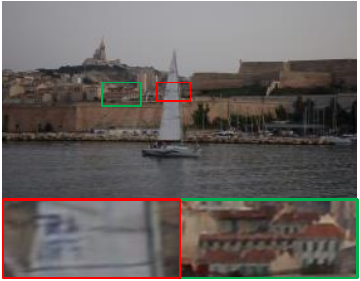}
		\label{Gao7}}
	\subfloat[MIMO-UNet+\cite{cho2021rethinking}]{\includegraphics[width=1.25in]{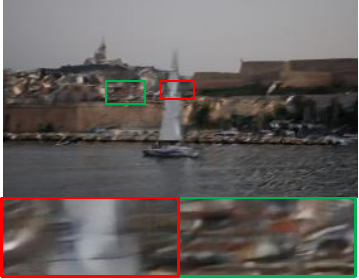}
		\label{MIMO+7}}
	\subfloat[Our CDCN]{\includegraphics[width=1.25in]{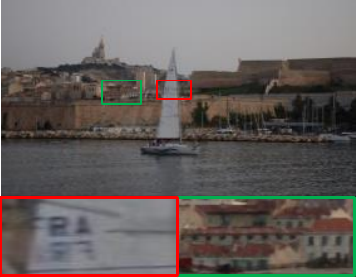}
		\label{CDCN7}}
	\caption{The qualitative evaluation comparisons of all the methods on the real blurred images.}
	\label{RealBlur comparisons1}
\end{figure*}

\begin{figure*}[!t]
	\centering
	\subfloat[Blur image]{\includegraphics[width=1.25in]{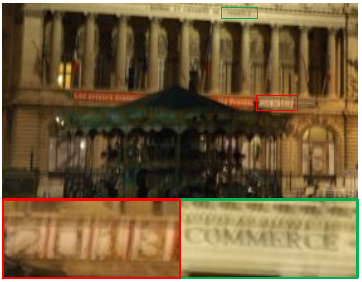}
		\label{Blur8}}
	\subfloat[SRN\cite{tao2018scale}]{\includegraphics[width=1.25in]{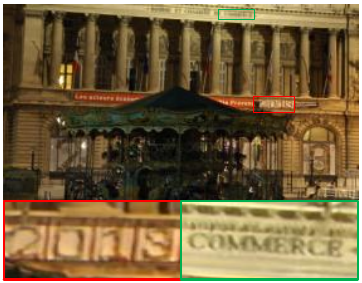}
		\label{SRN8}}
	\subfloat[Gao et al.\cite{gao2019dynamic}]{\includegraphics[width=1.25in]{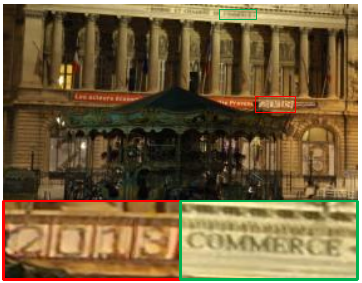}
		\label{Gao8}}
	\subfloat[MIMO-UNet+\cite{cho2021rethinking}]{\includegraphics[width=1.25in]{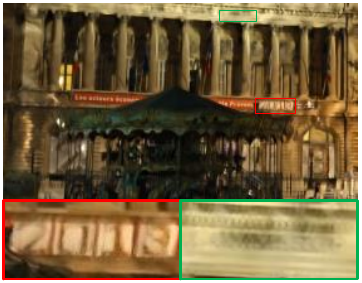}
		\label{MIMO+8}}
	\subfloat[Our CDCN]{\includegraphics[width=1.25in]{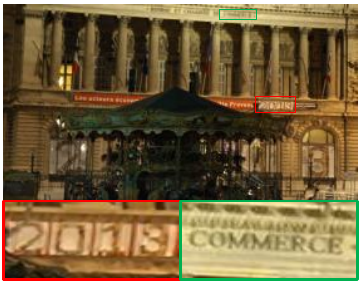}
		\label{CDCN8}}
	\caption{The qualitative evaluation comparisons of all the methods on the real blurred images.}
	\label{RealBlur comparisons2}
\end{figure*}

With the exception of the synthetic blurred images, we also use the real blurred images to further demonstrate the effectiveness of our method. For real blurred images, we compare our method with the methods \cite{tao2018scale}, \cite{gao2019dynamic} and \cite{cho2021rethinking}, and all methods are still trained on the 2103 GoPro training image pairs. Fig. \ref{RealBlur comparisons1} and Fig. \ref{RealBlur comparisons2} illustrate the qualitative evaluation comparisons of the models \cite{tao2018scale}, \cite{gao2019dynamic}, \cite{cho2021rethinking} and our CDCN model on two real blurred images. From Fig. \ref{RealBlur comparisons1} and Fig. \ref{RealBlur comparisons2} we can see that, methods \cite{tao2018scale}, \cite{gao2019dynamic} and \cite{cho2021rethinking} can not recover the real blurred image well, and there are too much flaws: the ringing artifacts, the blur, the distortion and the deformation, in the deblurred images. By contrast, our CDCN model can obtain the higher quality restoration images: less blur, less deformation, hardly  distortion and ringing artifacts, and sharper edges and more details Please see the Figs. \ref{SRN7}-\ref{CDCN7} and the Figs. \ref{SRN8}-\ref{CDCN8}, and the corresponding zoomed in regions. Figs. \ref{RealBlur comparisons1} and \ref{RealBlur comparisons2} demonstrate that, for the real blurred images, our method still can achieve higher quality restoration results. More experimental results for real blurred images can be available at https://github.com/wuyang1002431655/CDCN.

\section{CONCLUSION}

In this paper, we propose a novel constrained deformable convolutional network (CDCN) for accurate spatially-variant motion blur kernels estimation and high quality image restoration. Inspired by the PMPB model and the deformable convolution, a novel CDCR strategy is proposed to achieve accurate spatially-variant motion blur kernels estimation from only one single motion blurred image without the optical flow by proposing a PMPB-based reblurring loss function, which can make the learned sampling points fit the trajectory of the relative motion of each pixel well. Then, a novel MSML-MIMO encoder-decoder architecture is constructed, which possesses more powerful features extraction ability by utilizing and fusing of more information flows and informative features. Extensive experiments on both the synthetic benchmark datasets and the real blurred images show that our method can produce better deblurring results than the state-of-the-art SIDSBD methods in terms of both qualitative evaluation and quantitative metrics. Researching more general and more powerful constraint terms and incorporating them into the PMPB-based reblurring loss function for more accurate blur kernels estimation, and extending our CDCN to other types of blurred images (e.g. the defocus blurred images) are our future works.

\section{ACKNOWLEDGMENT}

The authors will thank the editor and all reviewers, and in addition, the authors thank Xu et al, Nah et al, Tao et al, Gao et al, Kupyn et al, Cho et al and so on for the source code or model provided.

\bibliographystyle{IEEEtran}
\bibliography{reference}

\begin{IEEEbiography}[{\includegraphics[width=1in,height=1.25in,clip,keepaspectratio]{ 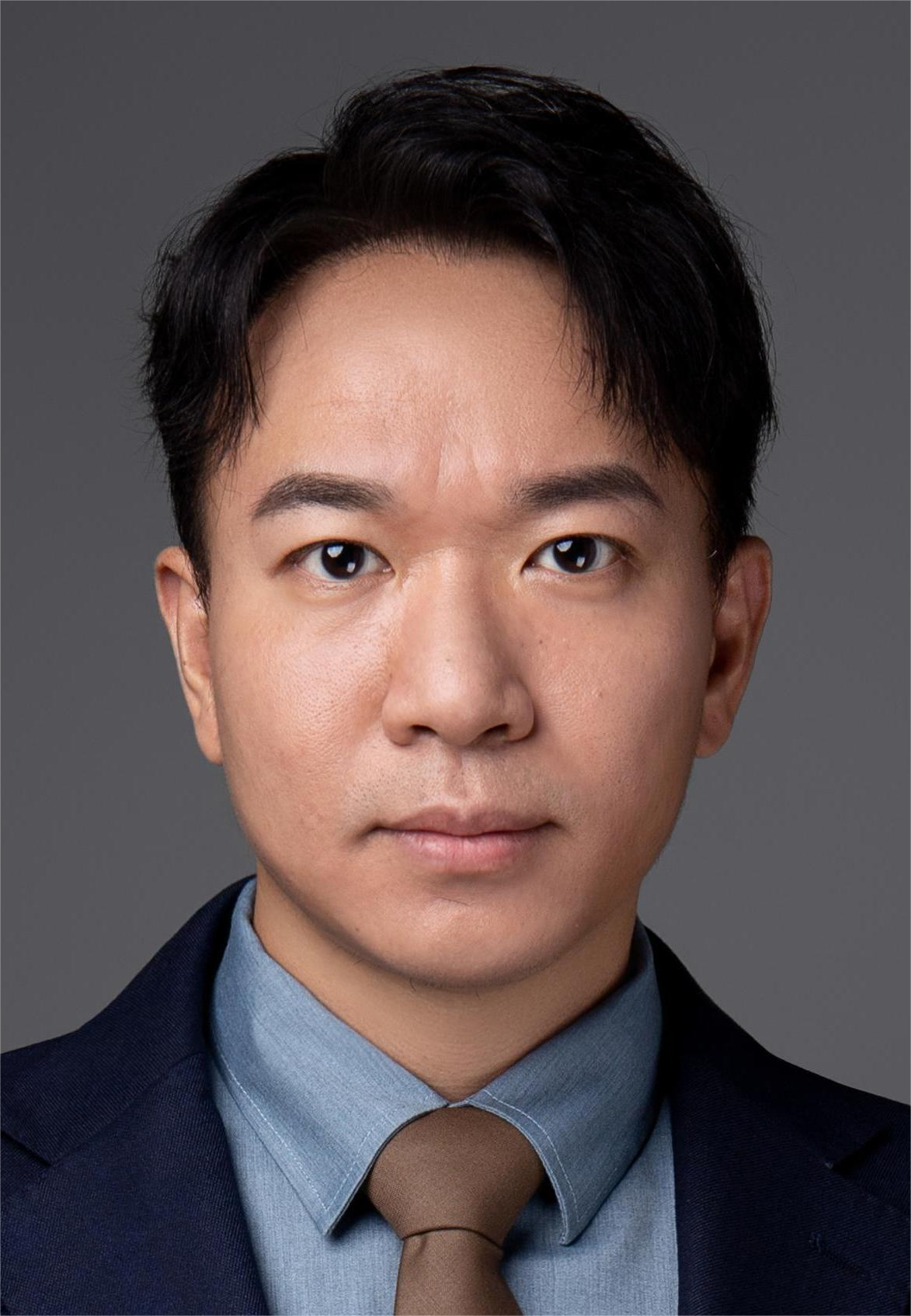}}]{Shu Tang}
 received the M.E. degree from Chongqing University of Posts and Telecommunications, Chongqing, China, in 2007, and the Ph.D. degree in Chongqing university, China, in 2013. He is currently an associate professor of the College of Computer Science and Technology at Chongqing University of Posts and Telecommunications, China. His research interests include signal processing, image processing, and computer vision. Email: tang- shu@cqupt.edu.cn.
\end{IEEEbiography}

\begin{IEEEbiography}[{\includegraphics[width=1in,height=1.25in,clip,keepaspectratio]{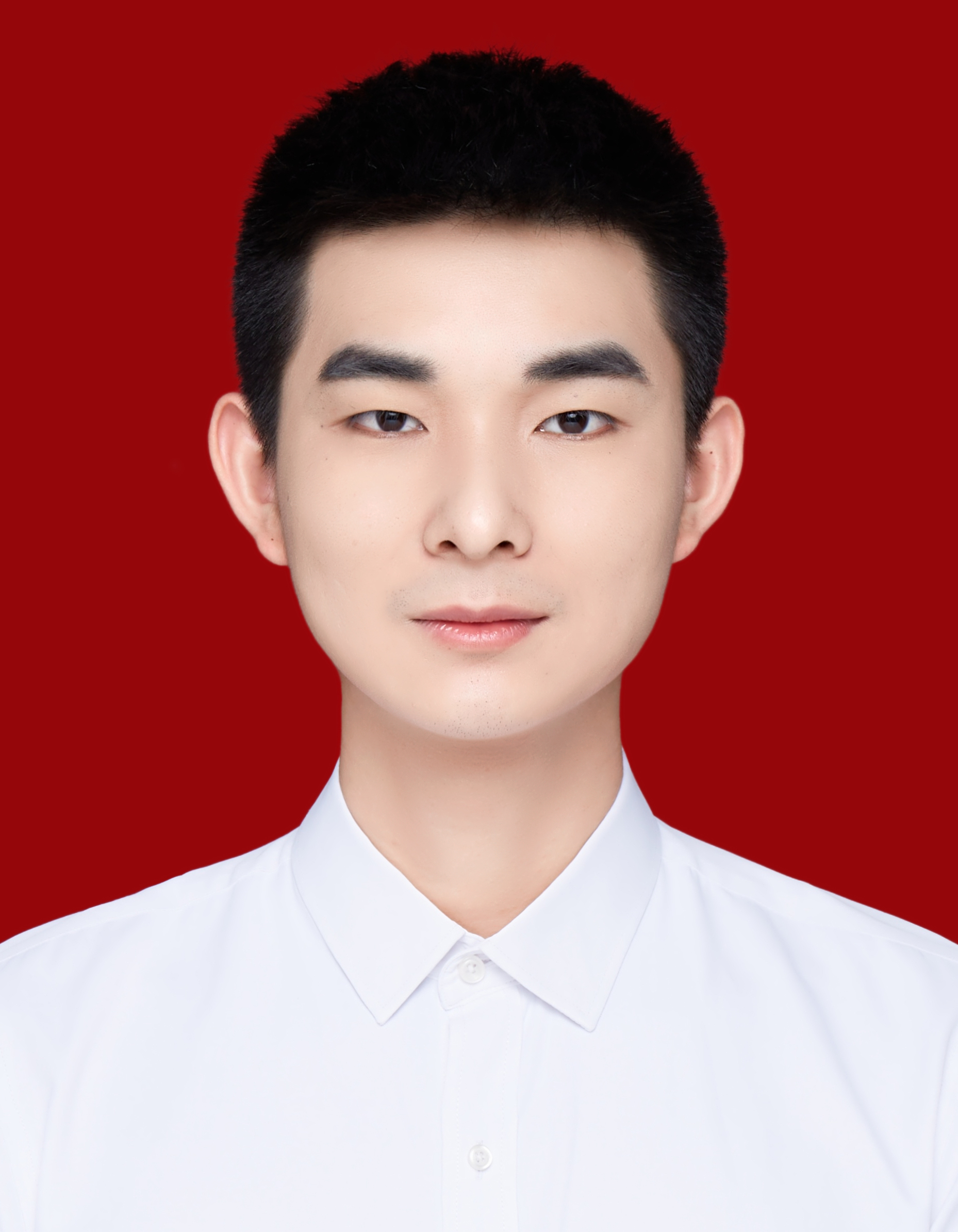}}]{Yang Wu}
	received his bachelor of engineering from Nanyang Institute of Technology, China, in 2019. He is studying for a master’s degree at Chongqing University of Posts and Telecommunications, China. His research interests include computer vision and deep learning. Email: 1002431655@qq.com.
\end{IEEEbiography}

\begin{IEEEbiography}[{\includegraphics[width=1in,height=1.25in,clip,keepaspectratio]{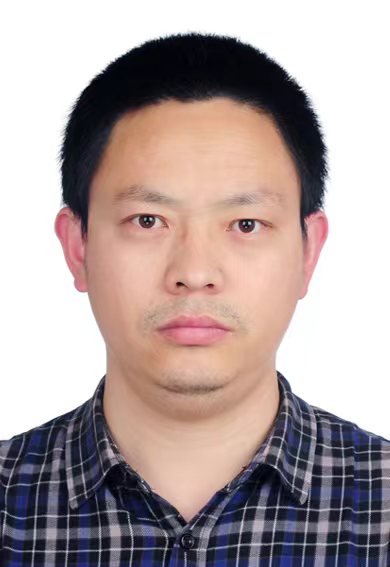}}]{Hongxing Qin}
	is now a Professor at Chongqing University, Chongqing. He received his PhD degree in pattern recognition from Shanghai Jiaotong University, in 2008. He worked as a postdoctoral researcher at Rutgers, the State University of New Jersey, from 2008 to 2009. His research interests include computer graphics, digital geometry processing, medical image processing, and visualization.\end{IEEEbiography}

\begin{IEEEbiography}[{\includegraphics[width=1in,height=1.25in,clip,keepaspectratio]{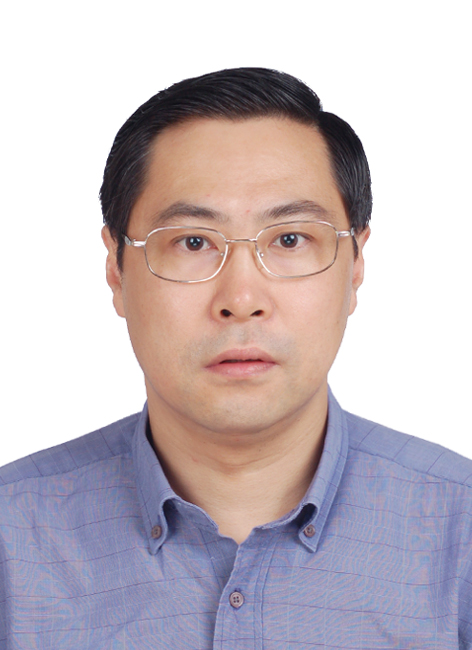}}]{Xianzhong Xie}
	 born in 1966, received his Ph.D. degree in communication and information systems from Xidian University, China, in 2000. He is currently the professor and Director of Chongqing Key Lab of Computer Network and Communication Technology at Chongqing University of Posts and Telecommunications, China. His research interests include MIMO precoding, cognitive radio networks, and cooperative communications. He is the principal author of five books on cooperative communications, 3G, MIMO, cognitive radio, and TDD technology. He has published more than 100 papers in journals and 30 papers in international conferences. Email: xiexzh@cqupt.edu.cn.\end{IEEEbiography}

\begin{IEEEbiography}[{\includegraphics[width=1in,height=1.25in,clip,keepaspectratio]{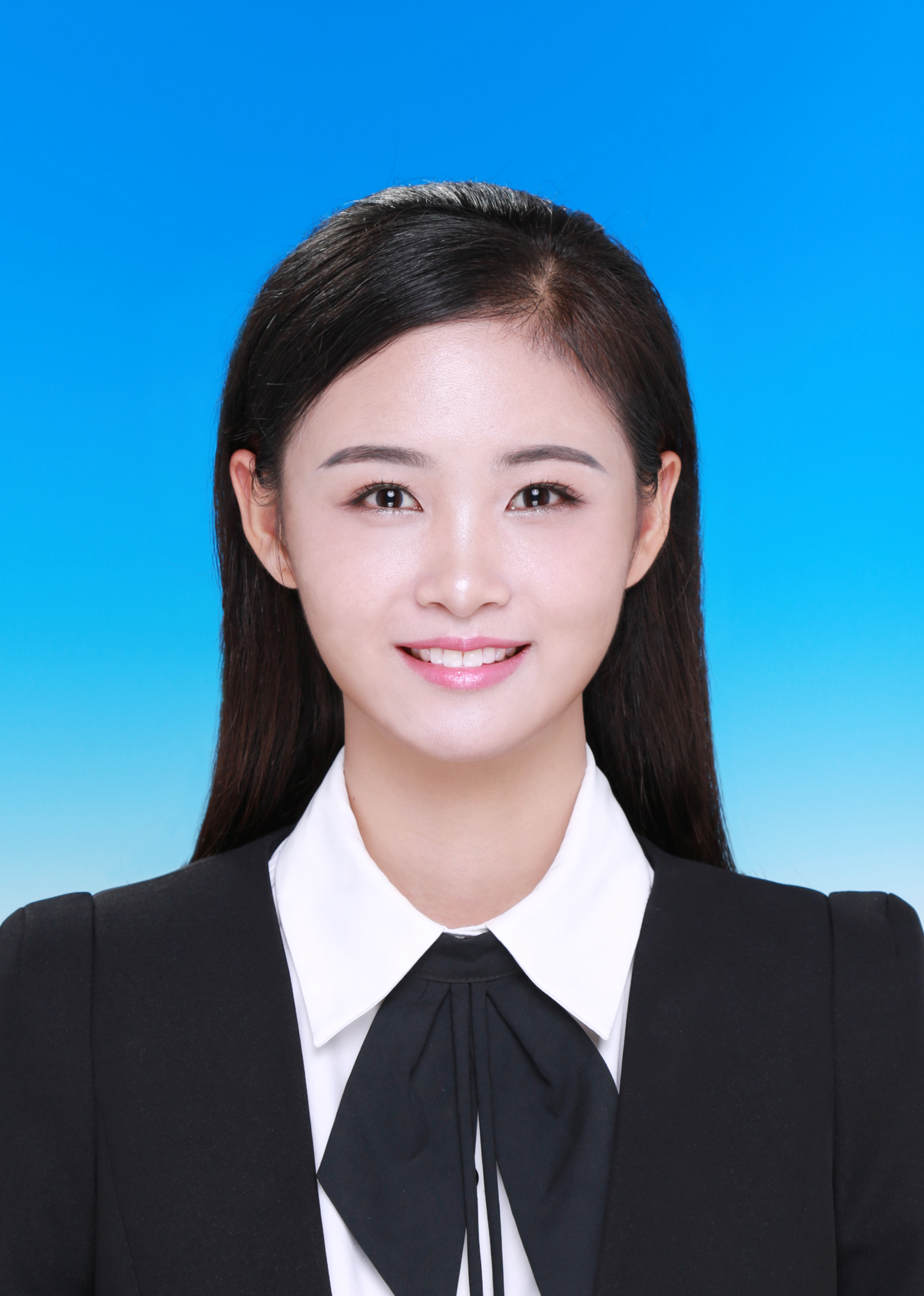}}]{Shuli Yang}
	is studying for her doctorate at Chongqing University of Posts and Telecommunications, China. Hers research interests include image super-resolution reconstruction and deep learning. Email: hiphop\_yang@163.com.
	\end{IEEEbiography}

\begin{IEEEbiography}[{\includegraphics[width=1in,height=1.25in,clip,keepaspectratio]{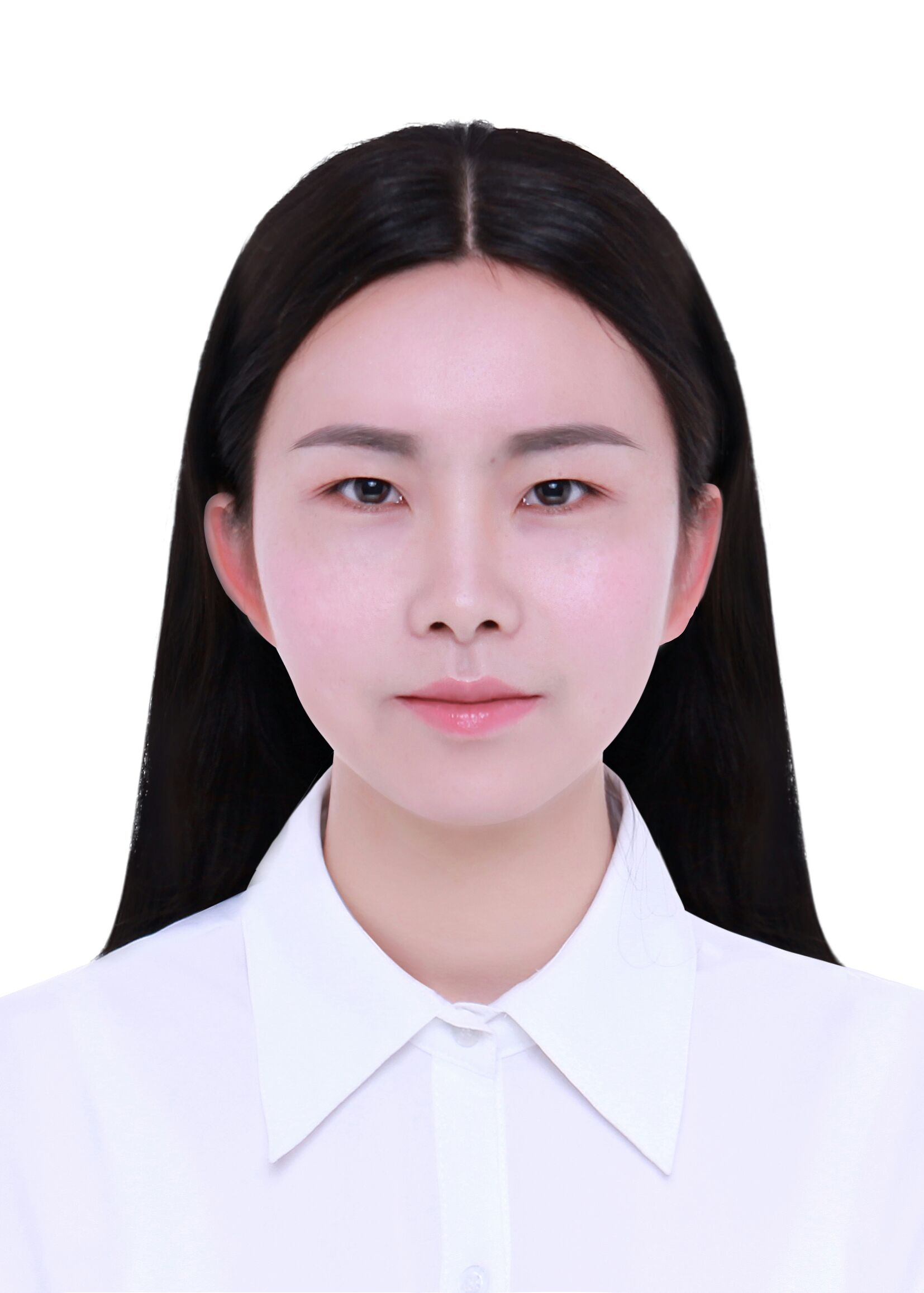}}]{Jing Wang}
	 received the bachelor’s degree in engineering from the Chongqing University of Posts and Telecommunications, Chongqing, China, in 2019, where she is currently studying with the School of Computer Science and Technology. Her research interest is image deblurring.
\end{IEEEbiography}

\end{document}